\documentclass[12pt]{article}
\usepackage{amsmath}
\usepackage{graphicx,psfrag,epsf}
\usepackage{enumerate}
\usepackage{natbib}
\usepackage{url} %
\usepackage{setspace}

\newcommand{\blind}{0}

\usepackage{amssymb}
\usepackage{amsthm}

\usepackage{xr}
\usepackage[top=0.85in,left=1.0in,right=1.0in,footskip=0.75in]{geometry}

\usepackage{changepage}

\usepackage[utf8]{inputenc}

\usepackage{textcomp,marvosym}

\usepackage[ruled]{algorithm}
\usepackage[noend]{algorithmic}

\usepackage{cite}

\usepackage{nameref,hyperref}

\usepackage{microtype}
\DisableLigatures[f]{encoding = *, family = * }

\usepackage{pifont}
\usepackage{color}

\newtheorem{theorem}{Theorem}
\newtheorem{definition}{Definition}
\newtheorem{condition}{Condition}
\newtheorem{lemma}{Lemma}

\DeclareMathOperator*{\argmin}{arg\,min}

\begin{document}

\def\spacingset#1{\renewcommand{\baselinestretch}%
{#1}\small\normalsize} \spacingset{1}

\if0\blind
{
  \title{\bf Gradient-based Regularization Parameter Selection for Problems with Non-smooth Penalty Functions}
  \author{Jean Feng\thanks{
    Jean Feng was supported by NIH grants DP5OD019820 and T32CA206089.
    Noah Simon was supported by NIH grant DP5OD019820.
    The content is solely the responsibility of the authors and does not necessarily represent the official views of the National Institutes of Health.}\\
    Department of Biostatistics, University of Washington\\
    and \\
    Noah Simon \\
    Department of Biostatistics, University of Washington}
  \maketitle
} \fi

\if1\blind
{
  \bigskip
  \bigskip
  \bigskip
  \begin{center}
    {\LARGE\bf Gradient-based Regularization Parameter Selection for Problems with Non-smooth Penalty Functions}
\end{center}
  \medskip
} \fi

\bigskip
\begin{abstract}
In high-dimensional and/or non-parametric regression problems, regularization (or penalization) is used to control model complexity and induce desired structure. Each penalty has a weight parameter that indicates how strongly the structure corresponding to that penalty should be enforced. Typically the parameters are chosen to minimize the error on a separate validation set using a simple grid search or a gradient-free optimization method. It is more efficient to tune parameters if the gradient can be determined, but this is often difficult for problems with non-smooth penalty functions. Here we show that for many penalized regression problems, the validation loss is actually smooth almost-everywhere with respect to the penalty parameters. We can therefore apply a modified gradient descent algorithm to tune parameters. Through simulation studies on example regression problems, we find that increasing the number of penalty parameters and tuning them using our method can decrease the generalization error.
\end{abstract}

\noindent%
\vfill

\newpage
\spacingset{1.45}
\section{Introduction}
Consider the usual regression framework with $p$ features, $\boldsymbol x_i = (x_{i1},\ldots,x_{ip})^\top$, and a response $y_i$ measured on each of $i=1,\ldots,n$ observations. Let $\boldsymbol X$ denote the $n \times p$ design matrix and $\boldsymbol y$ the response vector. Our goal here is to characterize the conditional relationship between $\boldsymbol y$ and $\boldsymbol X$. In simple low-dimensional problems this is often done by constructing an $f$ in some pre-specified class $\mathcal{F}$ that minimizes a measure of discrepancy between $\boldsymbol y$ and $f(\boldsymbol X)$. Generally, this discrepancy is quantified with some pre-specified loss, $L$. Often $\mathcal{F}$ will endow $f$ with some simple form (e.g. a linear function). For ill-posed or high-dimensional problems ($p \gg n$), there can often be an infinite number of solutions that minimize the loss function $L$ but have high generalization error. A common solution is to use regularization, or penalization, to select models with desirable properties, such as smoothness and sparsity.

In recent years, there has been much interest in combining regularization methods to produce models with multiple desired characteristics. For example, the elastic net \citep{zou2003regression} combines the lasso and ridge penalties; and the sparse group lasso \citep{simon2013sparse} combines the group lasso and lasso penalties. In Bayesian regression, a popular method for pruning irrelevant features is to use automatic relevance determination, which associates each feature with a separate regularization parameter \citep{neal1996bayesian}. From a theoretical viewpoint, multiple regularization parameters are required in certain cases to achieve oracle convergence rates. \citet{van2014additive} showed that when fitting additive models with varying levels of smoothness, the penalty parameter should be smaller for more ``wiggly" functions and vice versa. The general form of these regression problems is:
\begin{equation} \label {eq:basic}
\hat f(\boldsymbol{\lambda}) = \argmin_{f\in\mathcal{F}} L\left (\boldsymbol{y}, f (\boldsymbol{X}) \right ) + \sum\limits_{i=1}^J \lambda_i P_i(f)
\end{equation}
where $\{P_i\}_{i=1, ..., J}$ are the penalty functions and $\boldsymbol{\lambda} = (\lambda_1, \ldots, \lambda_J)^\top$ are the regularization parameters. 

Regularization parameters control the degree of various facets of model complexity, such as the amount of sparsity or smoothness. Often the goal is to set the parameters to minimize the fitted model's generalization error. One usually estimates this using a training/validation approach (or cross validation). In this approach, one fits a model on a training set $(\boldsymbol X_T, \boldsymbol y_T)$ and measures the model's error on a validation set $(\boldsymbol X_V, \boldsymbol y_V)$. The goal then is to choose penalty parameters $\boldsymbol{\lambda}$ that minimize the validation error, as formulated in the following joint optimization problem:
\begin{equation}
\begin{array}{c}
\min_{\boldsymbol{\lambda} \in \Lambda} L\left (\boldsymbol{y}_V, \hat f (\boldsymbol{X}_V | \boldsymbol{\lambda}) \right) \\
\text{s.t. } \hat f(\cdot | \boldsymbol{\lambda}) = \argmin_{f\in\mathcal{F}} L \left (\boldsymbol{y}_T, f (\boldsymbol{X}_T) \right) + \sum\limits_{i=1}^J \lambda_i P_i(f)
\end{array}
\label{jointopt}
\end{equation}
Here $\Lambda$ is some set that $\boldsymbol{\lambda}$ are known to be in, which is often just $\mathbb{R}^{J}_+$. We will refer to finding $\hat f (\cdot | \boldsymbol{\lambda})$ as solving the inner optimization problem.

The simplest approach to solving \eqref{jointopt} is brute force: one fits models over a grid of parameter values and selects the model with the lowest validation error. As long as the grid is large and fine enough, this method of ``grid search" will find a solution close to the global optimum. Unfortunately, it is computationally intractable in cases with more than two parameters since the runtime is exponential in the number of parameters.

More efficient methods treat \eqref{jointopt} as a continuous optimization problem, usually through a gradient-free or gradient-based approach. Gradient-free approaches include the Nelder-Mead simplex algorithm \citep{nelder1965simplex} and Bayesian optimization \citep{snoek2012practical, bergstra2011algorithms, hutter2011sequential}. Although Bayesian optimization is currently the gold standard in machine learning, gradient-free methods are generally unable to tune more than twenty or so parameters whereas gradient-based methods can handle hundreds or even thousands of parameters. To calculate the gradient, one can use reverse-mode differentiation through the entire training procedure \citep{maclaurin2015gradient} or implicit differentiation of the KKT conditions \citep{larsen1998adaptive, bengio2000gradient, foo2008efficient, lorbert2010descent}. Existing implicit differentiation methods all require the optimization criterion to be smooth. In this paper we show that many problems for which the inner optimization problem is non-smooth can be reformulated in a way that makes them amenable to tuning parameter optimization via gradient descent.

In Section~\ref{defineDescJointOpt}, we show that for certain joint optimization problems with non-smooth penalties, the outer optimization problem is still smooth almost everywhere. By locally reformulating the problem, we can apply the same implicit differentiation trick to obtain the gradient of the validation loss with respect to the penalty parameters. A descent-based algorithm is then proposed for tuning the penalty parameters. Section~\ref{sec:results} presents simulation studies comparing our method to gradient-free methods on regression problems with two to a hundred penalty parameters. Section~\ref{realDataResults} applies our method to gene expression data to predict colitis status.

\section{Gradient-based Joint Optimization}\label{defineDescJointOpt}
\subsection{Definition}
In this manuscript we will restrict ourselves to classes $\mathcal{F} = \left\{f_{\boldsymbol \theta}\middle| \boldsymbol \theta\in\Theta\right\}$, which, for a fixed sample size $n$, are in some finite dimensional space $\Theta$. This is not a large restriction: the class of linear functions meets this requirement; as does any class of finite dimensional parametric functions. Even non-parametric methods generally either use a growing basis expansion (e.g. Polynomial regression, smoothing-splines, wavelet-based-regression, locally-adaptive regression splines \citep{tsybakov2008introduction, wahba1981spline, donoho1994ideal, mammen1997locally}), or only evaluate the function at the observed data-points (eg. trend filtering, fused lasso, \citep{kim2009ell_1, tibshirani2005sparsity}). In these non-parametric problems, for any fixed $n$, $\mathcal{F}$ is representable as a finite dimensional class.
We can therefore rewrite \eqref{eq:basic} in the following form:
\begin{equation}\label{eq:train_disc}
\argmin_{\boldsymbol \theta \in \Theta} L(\boldsymbol{y}, f_{\boldsymbol \theta}(\boldsymbol{X})) + \sum\limits_{i=1}^J \lambda_i P_i(\boldsymbol \theta)
\end{equation}

Suppose that we use a training/validation split to select penalty parameters $\boldsymbol{\lambda} = (\lambda_1, ..., \lambda_J)^\top$. Let the data be partitioned into a training set $(\boldsymbol{y}_T , \boldsymbol{X}_T)$ and validation set $(\boldsymbol{y}_V, \boldsymbol{X}_V)$. We can rewrite the joint optimization problem \eqref{jointopt} over this finite-dimensional class as:
\begin{equation}
\begin{array}{c}
\argmin_{\boldsymbol{\lambda} \in \Lambda} L(\boldsymbol{y}_V, f_{\hat{\boldsymbol \theta}(\boldsymbol{\lambda})}(\boldsymbol{X}_V)) \\
\text{s.t. } {\hat{\boldsymbol \theta}(\boldsymbol{\lambda})} = \argmin_{\boldsymbol \theta \in \Theta} L(\boldsymbol{y}_T, f_{\boldsymbol \theta} (\boldsymbol{X}_T)) + \sum\limits_{i=1}^J \lambda_i P_i(\boldsymbol \theta)
\end{array}
\label{jointopt2}
\end{equation}
Note that joint optimization for $K$-fold cross validation is very similar. The outer criterion is the average validation loss over the models trained from all $K$ folds. See the Appendix for the full formulation.

For the remainder of the manuscript we will assume that the training criterion \eqref{eq:train_disc} is convex and has a unique minimizer. Also, we will assume that $L \left( \boldsymbol{y}_V, f_{\boldsymbol \theta}(\boldsymbol{X}_V) \right)$ is differentiable in $\boldsymbol \theta$. This assumption is met if both 1) $f_{\boldsymbol \theta}(\boldsymbol{X}_V)$ is continuous as a function of $\boldsymbol \theta$; and 2) $L\left(\boldsymbol{y}_V,\cdot\right)$ is smooth. Examples include the squared-error, logistic, and Poisson loss functions, though not the hinge loss.

\subsection{Smooth Training Criterion}
Here we present a brief summary of how to apply gradient descent when the training criterion is smooth. For more details, refer to \citet{bengio2000gradient}. Let the training criterion be denoted as
\begin{equation}
L_T\left(\boldsymbol \theta, \boldsymbol{\lambda}\right) \equiv L(\boldsymbol{y}_T, f_{\boldsymbol \theta} (\boldsymbol{X}_T)) + \sum\limits_{i=1}^J \lambda_i P_i(\boldsymbol \theta)
\label{train}
\end{equation}
To calculate the gradient, apply the chain rule
\begin{equation}
\nabla_{\boldsymbol{\lambda}} L \left( \boldsymbol{y}_V, f_{\hat{\boldsymbol \theta}(\boldsymbol{\lambda})}(\boldsymbol{X}_V) \right ) = 
\left [
\left . \frac{\partial}{\partial \boldsymbol \theta} L ( \boldsymbol{y}_V, f_{\boldsymbol \theta}(\boldsymbol{X}_V)) \right |_{\boldsymbol \theta=\hat{\boldsymbol \theta}(\boldsymbol \lambda)}
\right ]^\top 
\frac{\partial}{\partial \boldsymbol{\lambda}} \hat{\boldsymbol \theta}(\boldsymbol{\lambda})
\label{chainrule}
\end{equation}
The first term, $\frac{\partial}{\partial \boldsymbol \theta} L ( \boldsymbol{y}_V, f_{\boldsymbol \theta}(\boldsymbol{X}_V))$, is problem specific, but generally straightforward to calculate. To calculate the second term, $\frac{\partial}{\partial \boldsymbol{\lambda}} \hat{\boldsymbol \theta}(\boldsymbol{\lambda})$, we note that $\hat{\boldsymbol \theta}(\boldsymbol{\lambda})$ minimizes \eqref{train}. Since \eqref{train} is smooth,
\begin{equation}
\nabla_\theta 
L_T(\boldsymbol \theta, \boldsymbol{\lambda})
|_{\boldsymbol \theta = \hat {\boldsymbol \theta}(\boldsymbol{\lambda})}
= \boldsymbol{0}.
\label{eq:grad}
\end{equation}
Taking the derivative of both sides of \eqref{eq:grad} in $\boldsymbol{\lambda}$ and solving for $\frac{\partial}{\partial \boldsymbol{\lambda}} \hat{\boldsymbol \theta}(\boldsymbol{\lambda})$, we get:
\begin{equation}
\frac{\partial}{\partial \boldsymbol{\lambda}} \hat{\boldsymbol \theta}(\boldsymbol{\lambda}) = 
- \left . \left [
 \nabla_\theta^2 L_T (\boldsymbol \theta, \boldsymbol{\lambda} )^{-1}
\nabla_{\boldsymbol \theta} P(\boldsymbol \theta)
\right ]
\right |_{\boldsymbol \theta = \hat {\boldsymbol \theta}(\boldsymbol{\lambda})}
\label{implicitDifferentiation}
\end{equation}
where $\nabla_{\boldsymbol \theta} P(\boldsymbol \theta)$ is the matrix with columns $\{\nabla_{\boldsymbol \theta} P_i(\boldsymbol \theta)\}_{i=1:J}$.

We can plug \eqref{implicitDifferentiation} into \eqref{chainrule} to get $\nabla_{\boldsymbol{\lambda}} L \left ( \boldsymbol{y}_V, f_{\hat{\boldsymbol \theta}(\boldsymbol{\lambda})}(\boldsymbol{X}_V) \right )$. Note that because $\frac{\partial}{\partial \boldsymbol{\lambda}} \hat{\boldsymbol \theta}(\boldsymbol{\lambda})$ is defined in terms of $\hat{\boldsymbol \theta}\left(\boldsymbol{\lambda}\right)$, each gradient step requires minimizing the training criterion first. The gradient descent algorithm to solve \eqref{jointopt2} is given in Algorithm \ref{alg:smooth}.
\begin{algorithm}[H]
\caption{Gradient Descent for Smooth Training Criteria}
\label{alg:smooth}
\begin{spacing}{1}
\begin{algorithmic}
        \STATE{
	        	Initialize $\boldsymbol{\lambda}^{(0)}$.
	}
        \FOR{each iteration $k=0,1,...$ until stopping criteria is reached}
        \STATE{
        		Solve for $\hat {\boldsymbol \theta}(\boldsymbol{\lambda}^{(k)}) = \argmin_{\boldsymbol \theta \in \Theta} L_T(\boldsymbol \theta, \boldsymbol{\lambda}^{(k)})$.
	}
        \STATE{
        		Calculate the derivative of the model parameters with respect to the regularization parameters
                        	\begin{equation}
                        \frac{\partial}{\partial \boldsymbol{\lambda}} \hat{\boldsymbol \theta}(\boldsymbol{\lambda}) = 
                        - \left . \left [ \left (
                         \nabla_\theta^2 L_T(\boldsymbol \theta, \boldsymbol{\lambda}^{(k)})  \right )^{-1}
                        \nabla_{\boldsymbol \theta} P(\boldsymbol \theta)
                        \right ]
                        \right |_{\boldsymbol \theta = \hat {\boldsymbol \theta}(\boldsymbol{\lambda}^{(k)})}
                        \label{eq:gradient_hessian}
                        	\end{equation}
	}
	\STATE{
		Calculate the gradient
                      \begin{equation}
                      \left .
                      \nabla_{\boldsymbol{\lambda}} 
                      L \left (\boldsymbol{y_V}, f_{\hat {\boldsymbol \theta}(\boldsymbol{\lambda})}(\boldsymbol{X_V}) \right)
                      \right |_{\boldsymbol{\lambda} = \boldsymbol{\lambda}^{(k)}} =
                      \left (
                      \frac{\partial}{\partial \boldsymbol{\lambda}} \hat{\boldsymbol \theta}(\boldsymbol{\lambda}) \Big |_{\boldsymbol{\lambda}=\boldsymbol{\lambda}^{(k)}}
                      \right )^\top
                      \frac{\partial}{\partial \boldsymbol \theta} L(\boldsymbol{y}_V, f_{\boldsymbol \theta}(\boldsymbol{X_V})) \Big |_{\theta = \hat \theta(\boldsymbol{\lambda}^{(k)})} 
                      \end{equation}
        }
        \STATE{Perform gradient step with step size $t^{(k)}$
	\begin{equation}
	\boldsymbol{\lambda}^{(k+1)} := \boldsymbol{\lambda}^{(k)} -
	t^{(k)}
	\left . \nabla_{\boldsymbol{\lambda}} L \left( \boldsymbol{y}_V, f_{\hat{\theta}(\boldsymbol{\lambda})}(\boldsymbol{X}_V)  \right )
	\right |_{\boldsymbol{\lambda} = \boldsymbol{\lambda}^{(k)}}
	\end{equation}
	}
	\ENDFOR
\end{algorithmic}
\end{spacing}
\end{algorithm}

\subsection{Nonsmooth Training Criterion}
When the penalized training criterion in the joint optimization problem is not smooth, gradient descent cannot be directly applied. Nonetheless, we find that in many problems, the solution $\hat{\boldsymbol \theta}\left(\boldsymbol{\lambda}\right)$ is smooth at almost every $\boldsymbol{\lambda}$ (e.g. Lasso \citep{tibshirani1996regression}, Group Lasso \citep{yuan2006model}, Trend Filtering \citep{kim2009ell_1}); this means that we can indeed apply gradient descent in practice. In this section, we characterize these problems that are almost everywhere smooth. In addition, we provide a solution for deriving $\frac{\partial}{\partial \boldsymbol{\lambda}} \hat{\boldsymbol \theta}(\boldsymbol{\lambda})$ since calculating the gradient is a challenge in and of itself. This is then incorporated into an algorithm for tuning $\boldsymbol{\lambda}$ using gradient descent.

To characterize problems that are almost everywhere smooth, we begin with three definitions:
\begin{definition}
The differentiable space of a real-valued function $L$ at a point $\boldsymbol \eta$ in its domain is the set of vectors along which the directional derivative of $L$ exists.
\begin{equation}
\Omega^{L}(\boldsymbol \eta) = \left \{ \boldsymbol u \middle | \lim_{\epsilon \rightarrow 0} \frac{L(\boldsymbol \eta + \epsilon \boldsymbol u) - L(\boldsymbol \eta)}{\epsilon} \text{ exists } \right \}
\end{equation}
\end{definition}

\begin{definition}
$S$ is a local optimality space for a convex function $L(\cdot, \boldsymbol \lambda_0)$ if there exists a neighborhood $W$ containing $\boldsymbol \lambda_0$ such that for every $\boldsymbol \lambda \in W$,
\begin{equation}
\argmin_{\boldsymbol \theta \in \Theta} L(\boldsymbol \theta, \boldsymbol \lambda) =
\argmin_{\boldsymbol \theta \in S} L(\boldsymbol \theta, \boldsymbol \lambda)
\end{equation}
\end{definition}

\begin{definition}
Consider a real-valued function $f: \mathbb{R}^p \mapsto \mathbb{R}$. Let matrix $\boldsymbol U = [ \boldsymbol u_1 \hdots \boldsymbol u_q ] \in \mathbb{R}^{p \times q}$ have orthonormal columns. Suppose the first and second directional derivatives of $f$ with respect to the columns in $\boldsymbol U$ exist. The Gradient vector and Hessian matrix of $f$ with respect to $\boldsymbol U$ are defined respectively as
\begin{equation}\label{eq:hess}
_{\boldsymbol U} \nabla f  =
\left (
\begin{array}{c}
\frac{\partial f}{\partial  u_1} \\
\frac{\partial f}{\partial u_2} \\
\vdots\\
\frac{\partial f}{\partial u_q}\\
\end{array}
\right ) \in \mathbb{R}^q;
\quad
_{\boldsymbol U}\nabla^2 f =
\left (
\begin{array}{cccc}
\frac{\partial^2 f}{\partial u_1^2} & \frac{\partial^2 f}{\partial u_1 \partial u_2} & ...  & \frac{\partial^2 f}{\partial u_1 \partial u_q} \\
\frac{\partial^2 f}{\partial u_2 \partial u_1} & \frac{\partial^2 f}{\partial u_2^2} & ...  & \frac{\partial^2 f}{\partial u_2 \partial u_q} \\
\vdots & \vdots &  \ddots & \vdots \\
\frac{\partial^2 f}{\partial u_q \partial u_1} & \frac{\partial^2 f}{\partial u_q \partial u_2} & ...  & \frac{\partial^2 f}{\partial u_q^2} \\
\end{array}
\right ) \in \mathbb{R}^{q \times q}
\end{equation}
\end{definition}

Using these definitions we can now give three conditions which together are sufficient for the differentiability of $L \left( \boldsymbol{y}_V, f_{\hat{\boldsymbol \theta}(\boldsymbol{\lambda})}(\boldsymbol{X}_V) \right )$ almost everywhere.

\begin{condition}
For almost every $\boldsymbol{\lambda}$, the differentiable space $\Omega^{L_T(\cdot, \boldsymbol{\lambda})}(\hat{\boldsymbol \theta}\left(\boldsymbol{\lambda}\right))$ is a local optimality space for $L_T\left(\cdot,\boldsymbol{\lambda}\right)$.
\label{condn:local_is_diff}
\end{condition}

\begin{condition}
For almost every $\boldsymbol{\lambda}$, $L_T\left(\cdot, \cdot\right)$ restricted to $\Omega^{L_T(\cdot, \cdot)}(\hat{\boldsymbol \theta}\left(\boldsymbol{\lambda}\right), \boldsymbol{\lambda})$ is twice continuously differentiable within some neighborhood of $\boldsymbol{\lambda}$.
\label{condn:twice_diff}
\end{condition}

\begin{condition}
For almost every $\boldsymbol{\lambda}$, there exists an orthonormal basis $\boldsymbol U$ of $\Omega^{L_T(\cdot, \boldsymbol{\lambda})}(\hat{\boldsymbol \theta}\left(\boldsymbol{\lambda}\right))$ such that the Hessian of $L_T\left(\cdot, \boldsymbol{\lambda}\right)$ at $\hat{\boldsymbol \theta}\left(\boldsymbol{\lambda}\right)$ with respect to $\boldsymbol U$ is invertible.
\label{condn:hessian}
\end{condition}
Note that if condition 3 is satisfied, the Hessian of $L_T \left(\cdot, \boldsymbol{\lambda}\right)$ with respect to any orthonormal basis of $\Omega^{L_T(\cdot, \boldsymbol{\lambda})}(\hat{\boldsymbol \theta}\left(\boldsymbol{\lambda}\right))$ is invertible.

Putting all these conditions together, the following theorem establishes that the gradient exists almost everywhere and provides a recipe for calculating it.

\begin{theorem}
Suppose our optimization problem is of the form in \eqref{jointopt2}, with $L_T\left(\boldsymbol \theta, \boldsymbol{\lambda}\right)$ defined as in \eqref{train}. Suppose that $L \Big( \boldsymbol{y}_V, f_{\boldsymbol \theta}(\boldsymbol{X}_V)\Big)$ is continuously differentiable in $\boldsymbol \theta$, and conditions $1$, $2$, and $3$, defined above, hold. Then the validation loss $L(\boldsymbol{y_V}, f_{\hat {\boldsymbol\theta}(\boldsymbol{\lambda})}(\boldsymbol{X_V}))$ is continuously differentiable with respect to $\boldsymbol{\lambda}$ for almost every $\boldsymbol{\lambda}$. Furthermore, the gradient of $L(\boldsymbol{y_V}, f_{\hat \theta(\boldsymbol{\lambda})}(\boldsymbol{X_V}))$, where it is defined, is
\begin{equation}
\nabla_{\boldsymbol{\lambda}} L \left ( \boldsymbol{y_V}, f_{\hat {\boldsymbol \theta}(\boldsymbol{\lambda})}(\boldsymbol{X_V}) \right) =
\left [ \left .
\frac{\partial}{\partial \boldsymbol \theta} L(\boldsymbol{y_V}, f_{\boldsymbol \theta}(\boldsymbol{X_V}))
\right |_{\boldsymbol \theta=\tilde{\boldsymbol \theta}(\boldsymbol \lambda)} \right ]^\top
\frac{\partial}{\partial \boldsymbol{\lambda}} \tilde{\boldsymbol \theta}(\boldsymbol{\lambda})
\end{equation}
where
\begin{equation}
\tilde{\boldsymbol \theta}(\boldsymbol{\lambda}) = \argmin_{\boldsymbol \theta \in \Omega^{L_T(\cdot, \boldsymbol{\lambda})}(\hat {\boldsymbol \theta}(\boldsymbol{\lambda}))} L_T(\boldsymbol \theta , \boldsymbol{\lambda})
\label{restrictedmodelparams}
\end{equation}
\label{thethrm}
\end{theorem}

We can therefore construct a gradient descent procedure based on the model parameter constraint in \eqref{restrictedmodelparams}. At each iteration, let matrix $\boldsymbol U$ have orthonormal columns spanning the differentiable space $\Omega^{L_T(\cdot, \boldsymbol{\lambda})}(\hat {\boldsymbol \theta}(\boldsymbol{\lambda}))$. Since this space is also a local optimality space, it is sufficient to minimize the training criterion over the column space of $\boldsymbol U$. The joint optimization problem can be reformulated using $\boldsymbol{\theta} = \boldsymbol U \boldsymbol \beta$ as the model parameters instead:
\begin{equation}
\begin{array}{c}
\min_{\boldsymbol \lambda \in \Lambda} L(\boldsymbol y_V, f_{\boldsymbol U \hat{\boldsymbol \beta} (\boldsymbol \lambda) }(\boldsymbol X_V)) \\
\text{s.t. } \hat{\boldsymbol \beta} (\boldsymbol \lambda) =
\argmin_{\boldsymbol \beta}
L_T (\boldsymbol U \boldsymbol \beta, \boldsymbol{\lambda} )
\end{array}
\end{equation}

This locally equivalent problem now reduces to the simple case where the training criterion is smooth. Implicit differentiation on the gradient condition gives us $\frac{\partial}{\partial \boldsymbol \lambda}\hat{\boldsymbol \beta}(\boldsymbol \lambda)$ and, thereby,
$\frac{\partial}{\partial \boldsymbol \lambda}
\hat{\boldsymbol \theta}(\boldsymbol \lambda) =
\boldsymbol U
\frac{\partial}{\partial \boldsymbol \lambda}\hat{\boldsymbol \beta}(\boldsymbol \lambda)
$.
Note that because the differentiable space is a local optimality space and is thus locally constant, we can treat $\boldsymbol U$ as a constant in the gradient derivations. Algorithm~\ref{alg:gradDescent} provides the exact steps for tuning the regularization parameters.

\begin{algorithm}[t]
	\caption{\label{alg:gradDescent} Gradient-based Joint Optimization}
	\begin{spacing}{1.1}
		\begin{algorithmic}
			\STATE{
				Initialize $\boldsymbol{\lambda}^{(0)}$.
			}
			\FOR {each iteration $k=0,1,...$ until stopping criteria is reached}
			\STATE{
				Solve for $\hat {\boldsymbol \theta}(\boldsymbol{\lambda}^{(k)}) = \argmin_{\theta \in \Theta} L_T(\boldsymbol \theta, \boldsymbol{\lambda}^{(k)})$.
			}
			\STATE{
				Construct matrix $\boldsymbol U^{(k)}$, an orthonormal basis of $\Omega^{L_T(\cdot, \boldsymbol{\lambda})}\left (\hat{\boldsymbol \theta}(\boldsymbol{\lambda}^{(k)}) \right )$.
			}
			\STATE{
				Define the locally equivalent joint optimization problem
				\begin{equation*}
				\begin{array}{c}
				\min_{\boldsymbol \lambda \in \Lambda} L(\boldsymbol y_V, f_{\boldsymbol U^{(k)} \hat{\boldsymbol \beta} (\boldsymbol \lambda) }(\boldsymbol X_V)) \\
				\text{s.t. } \hat{\boldsymbol \beta} (\boldsymbol \lambda) =
				\argmin_{\boldsymbol \beta}
				L_T (\boldsymbol U^{(k)}\boldsymbol \beta, \boldsymbol{\lambda} )
				\end{array}
				\end{equation*}
			}              
			\STATE{
				Calculate $\frac{\partial}{\partial \boldsymbol \lambda} \hat{\boldsymbol \beta}(\boldsymbol{\lambda})|_{\boldsymbol{\lambda} = \boldsymbol{\lambda}^{(k)}}$ where 
				\begin{equation*}
				\frac{\partial}{\partial \boldsymbol \lambda} \hat{\boldsymbol \beta}(\boldsymbol{\lambda}) =
				-  \left . \left[
				\left (
				_{\boldsymbol U^{(k)}}\nabla^2 L_T (\boldsymbol U^{(k)}\boldsymbol \beta, \boldsymbol{\lambda} )
				\right )^{-1}
				{}_{\boldsymbol U^{(k)}}\nabla P(\boldsymbol U^{(k)}\boldsymbol \beta)
				\right ]
				\right |_{\boldsymbol \beta =  \hat{\boldsymbol \beta}(\boldsymbol \lambda)}
				\end{equation*}
			}
			
			\STATE{
				Calculate the gradient of the validation loss %
				\begin{equation*}
				\nabla_{\boldsymbol{\lambda}} L \left (\boldsymbol{y_V}, f_{\hat \theta(\boldsymbol{\lambda})}(\boldsymbol{X_V}) \right ) =
				\left [
				\boldsymbol U^{(k)}
				\frac{\partial}{\partial \boldsymbol \lambda} \hat{\boldsymbol \beta}(\boldsymbol{\lambda})
				\right ]^\top
				\left [ \left .
				_{\boldsymbol U^{(k)}}\nabla L\left (\boldsymbol{y_V}, f_{\boldsymbol U^{(k)}\boldsymbol \beta}(\boldsymbol{X_V}) \right )
				\right |_{\boldsymbol \beta = \hat{\boldsymbol \beta}(\boldsymbol \lambda)}
				\right ]
				\end{equation*}
			}
			\STATE{
				Perform the gradient update with step size $t^{(k)}$
				\begin{equation*}
				\boldsymbol{\lambda}^{(k+1)} :=
				\boldsymbol{\lambda}^{(k)} - t^{(k)}
				\left .
				\nabla_{\boldsymbol{\lambda}} L \left (\boldsymbol{y_V}, f_{\hat \theta(\boldsymbol{\lambda})}(\boldsymbol{X_V}) \right )
				\right |_{\boldsymbol{\lambda} = \boldsymbol{\lambda}^{(k)}}
				\end{equation*}
			}
			\ENDFOR
		\end{algorithmic}
	\end{spacing}
\end{algorithm}

\subsection{Examples}\label{sec:examples}

To better understand the proposed gradient descent procedure, we present example joint optimization problems with nonsmooth criteria and their corresponding gradient calculations.

For ease of notation, we let $S_{\boldsymbol{\lambda}}$ denote the differentiable space of $L_T(\cdot, \boldsymbol{\lambda})$ at $\hat{\boldsymbol{\theta}}(\boldsymbol{\lambda})$. For each of the example regressions, justification that the conditions in Theorem~\ref{thethrm} are satisfied is included in the Appendix. Note that in some examples below, we add a ridge penalty with a fixed small coefficient $\epsilon > 0$ to ensure that the problem satisfies Condition \ref{condn:hessian}. Intuitively, the additional ridge penalty makes the models more well-behaved. For example, in the elastic net, combining the ridge and lasso penalties results in models that often exhibit better properties than just the lasso alone.

\subsubsection{Elastic Net}\label{sec:enet}

The elastic net \citep{zou2003regression} is a linear combination of the lasso and ridge penalties that encourages both sparsity and grouping of predictors. We tune the regularization parameters $\boldsymbol{\lambda} = (\lambda_1, \lambda_2)^\top$ for the joint optimization problem:
\begin{equation}
\begin{array}{c}
\min_{\boldsymbol{\lambda} \in \mathbb{R}^2_{+}} \frac{1}{2} \| \boldsymbol{y}_V - \boldsymbol{X}_V \hat{\boldsymbol{\theta}} (\boldsymbol \lambda) \| ^2 \\
\text{s.t. }
\hat{\boldsymbol{\theta}} (\boldsymbol{\lambda}) = \argmin_{\boldsymbol{\theta}} \frac{1}{2} \| \boldsymbol{y}_T - \boldsymbol{X}_T \boldsymbol{\theta} \| ^2
+ \lambda_1 \| \boldsymbol{\theta} \|_1
+ \frac{1}{2}\lambda_2 \| \boldsymbol{\theta} \|_2^2
\end{array}
\end{equation}

The first step to finding the gradient is determining the differentiable space. Let the nonzero indices of $\hat{\boldsymbol{\theta}}(\boldsymbol{\lambda})$ be denoted $I(\boldsymbol\lambda) = \{i | \hat{\theta}_i(\boldsymbol\lambda) \ne 0 \text{ for } i=1,...,p \}$ and let $\boldsymbol I_{I(\boldsymbol \lambda)}$ be a submatrix of the identity matrix with columns $I(\boldsymbol\lambda)$. Since $|\cdot|$ is not differentiable at zero, the directional derivatives of $||\boldsymbol \theta||_1$ only exist along directions spanned by the columns of $\boldsymbol I_{I(\boldsymbol \lambda)}$. Thus the differentiable space at $\boldsymbol \lambda$ is
$
S_{\boldsymbol{\lambda}} = span(\boldsymbol I_{I(\boldsymbol \lambda)})
\label{eq:en_diff_space}
$.

Let $\boldsymbol{X}_{T, I(\boldsymbol\lambda)} = \boldsymbol{X}_T \boldsymbol{I}_{I(\boldsymbol \lambda)}$ and $\boldsymbol{X}_{V, I(\boldsymbol\lambda)}  = \boldsymbol{X}_V \boldsymbol{I}_{I(\boldsymbol \lambda)}$.  The locally equivalent joint optimization problem is
\begin{equation}
\begin{array}{c}
\min_{\boldsymbol{\lambda} \in \mathbb{R}^2_{+}} \frac{1}{2} \| \boldsymbol{y}_V - \boldsymbol{X}_{V, I(\boldsymbol \lambda)} \hat{\boldsymbol{\beta}} (\boldsymbol \lambda) \| ^2 \\
\text{s.t. }
\hat{\boldsymbol{\beta}} (\boldsymbol{\lambda}) = \argmin_{\boldsymbol \beta} \frac{1}{2} \| \boldsymbol{y}_T - \boldsymbol{X}_{T, I(\boldsymbol \lambda)} \boldsymbol \beta \| ^2
+ \lambda_1 \| \boldsymbol \beta \|_1
+ \frac{1}{2}\lambda_2 \| \boldsymbol \beta \|_2^2
\end{array}
\end{equation}

Since the problem is now smooth, we can apply the chain rule and \eqref{implicitDifferentiation} to get the gradient of the validation loss
\begin{equation}
\nabla_{\boldsymbol \lambda} L(\boldsymbol{y_V}, f_{\hat{\boldsymbol{\theta}}(\lambda)}(\boldsymbol{X_V})) =
- \left (
\boldsymbol{X}_{V, I(\boldsymbol\lambda)}
\frac{\partial}{\partial \boldsymbol \lambda} \hat{\boldsymbol{\beta}}(\boldsymbol{\lambda})
\right )^{\top}
\left (
\boldsymbol y_V - \boldsymbol{X}_{V, I(\boldsymbol\lambda)} \hat{\boldsymbol{\beta}} (\boldsymbol{\lambda})
\right )
\end{equation}
where
\begin{equation}
\frac{\partial}{\partial \boldsymbol \lambda} \hat{\boldsymbol{\beta}}(\boldsymbol{\lambda}) = 
- \left ( 
\boldsymbol{X}_{T, I(\boldsymbol\lambda)}^\top \boldsymbol{X}_{T, I(\boldsymbol\lambda)} + \lambda_2 \boldsymbol{I}
\right )^{-1}
\begin{bmatrix}
sgn \left (\hat{\boldsymbol{\beta}} (\boldsymbol{\lambda}) \right ) &
\hat{\boldsymbol{\beta}} (\boldsymbol{\lambda})
\end{bmatrix}
\end{equation}

\subsubsection{Additive Models with Sparsity and Smoothness Penalties}\label{sec:additive}

Now consider the nonparametric regression problem given response $y$ and covariates $\boldsymbol{x} \in \mathbb{R}^p$. We suppose $y$ is the sum of $p$ univariate functions:
\begin{equation}
y = \sum_{i=1}^p f_i(x_i) + \epsilon
\end{equation}
where $\epsilon$ are independent with mean zero.
Let $\boldsymbol{\theta}^{(i)} \equiv (f_i(x_{i1}), ..., f_i(x_{in}))$ be estimates of functions $f_i$ at the observations. The model is fit using the least squares loss with sparsity and smoothness penalties. More specifically, for each estimate $\boldsymbol{\theta}^{(i)}$, we add a group lasso penalty $\| \cdot \|_2$ to encourage sparsity at the function level and a lasso penalty on the second-order discrete differences to encourage smooth function estimates. Hence there are a total of $2p$ non-smooth functions in the training criterion.

This regression problem usually employs two regularization parameters, one for the sum of the sparsity penalties and one for the sum of the smoothness penalties \citep{buhlmann2011statistics}. In the joint optimization problem below, we use a separate penalty parameter for each of the smoothness penalties, resulting in a total of $p+1$ penalty parameters. Ideally the parameters are tuned such that functions with nearly constant slope have large penalty parameters and ``wiggly" functions have small penalty parameters.

Define matrices $\boldsymbol{I}_T$ and $\boldsymbol{I}_V$ such that $\boldsymbol I_T \boldsymbol{\theta}^{(i)}$ and $\boldsymbol I_V \boldsymbol{\theta}^{(i)}$ are estimates for $f_i$ at the training and validation inputs, respectively. The joint optimization problem is
\begin{equation}
\begin{array}{c}
\min_{\boldsymbol\lambda \in \mathbb{R}^{p+1}_{+}} \frac{1}{2}
\left \|
\boldsymbol{y}_V
- \boldsymbol{I}_V \sum_{i=1}^p \hat{\boldsymbol{\theta}}^{(i)}(\boldsymbol{\lambda})
\right \|^2_2 \\
\text{s.t. }
\hat{\boldsymbol{\theta}}(\boldsymbol{\lambda}) =
\argmin_{\boldsymbol{\theta}}
\frac{1}{2} \left \|
\boldsymbol{y}_T
- \boldsymbol{I}_T \sum_{i=1}^p \boldsymbol{\theta}^{(i)} \right \|^2_2
+ \lambda_0 \sum_{i=1}^p \| \boldsymbol{\theta}^{(i)} \|_2
+ \sum_{i=1}^p \lambda_i \left \| \boldsymbol{D}^{(2)}_{\boldsymbol{x}_i} \boldsymbol{\theta}^{(i)} \right \|_1
+ \frac{\epsilon}{2}  \sum_{i=1}^p \| \boldsymbol{\theta}^{(i)} \|_2^2
\end{array}
\label{aplmProblem}
\end{equation}

The differentiable space is straightforward to determine for this problem, though requires bulky notation. Define $I_i(\boldsymbol{\lambda})$ for $i=1,...,p$ to be the indices along which smoothness penalty is not differentiable
\begin{equation}
I_i(\boldsymbol{\lambda}) = \left \{j | \left (\boldsymbol{D}^{(2)}_{\boldsymbol{x}_i} \hat{\boldsymbol \theta}^{(i)}(\boldsymbol\lambda) \right )_j = 0 \text{ for } j=1,...,n-2 \right \}
\end{equation}
and
\begin{equation}
J(\boldsymbol{\lambda}) = \left \{ i | \hat{\boldsymbol \theta}^{(i)}(\boldsymbol\lambda) \ne \boldsymbol 0 \text{ for } i=1,...,p \right \}
\end{equation}
The group lasso penalty $\|\cdot\|_2$ is not differentiable in any direction at ${\bf 0}$ and is differentiable in all directions elsewhere. Then $S_{\boldsymbol \lambda} = span(\boldsymbol {U}^{(1)}) \oplus ... \oplus span(\boldsymbol {U}^{(p)}) $ where $\boldsymbol {U}^{(i)} = \boldsymbol{0}$ if $\hat{\boldsymbol \theta}^{(i)}(\boldsymbol\lambda) = \boldsymbol 0$ and $\boldsymbol {U}^{(i)}$ is an orthonormal basis of $\mathcal{N}(\boldsymbol{I}_{I_i(\lambda)}\boldsymbol{D}^{(2)}_{\boldsymbol{x}_i})$ otherwise.

Now we can define the locally equivalent joint optimization problem:
\begin{equation}
\begin{array}{c}
\min_{\boldsymbol{\lambda} \in \mathbb{R}^{p+1}_{+}}
\frac{1}{2} \left \| \boldsymbol{y}_V
- \boldsymbol{I}_V \sum_{i\in J(\boldsymbol\lambda)} \boldsymbol {U}^{(i)} \hat{\boldsymbol{\beta}}^{(i)}(\boldsymbol{\lambda})
 \right \|^2_2 \\
\text{s.t. }
\hat{\boldsymbol{\beta}}(\boldsymbol{\lambda}) = \argmin_{\boldsymbol \beta}
\frac{1}{2} \left \| \boldsymbol{y}_T
- \boldsymbol{I}_T \sum_{i\in J(\boldsymbol\lambda)} \boldsymbol {U}^{(i)} \boldsymbol{\beta}^{(i)}
\right \|^2_2
+ \lambda_0 \sum_{i\in J(\boldsymbol\lambda)}  \| \boldsymbol {U}^{(i)} \boldsymbol{\beta}^{(i)} \|_2 \\
+ \sum_{i\in J(\boldsymbol\lambda)} \lambda_i \left \| \boldsymbol{D}^{(2)}_{\boldsymbol{x}_i} \boldsymbol {U}^{(i)} \boldsymbol{\beta}^{(i)} \right \|_1
+ \frac{\epsilon}{2}  \sum_{i\in J(\boldsymbol\lambda)} \| \boldsymbol {U}^{(i)} \boldsymbol{\beta}^{(i)} \|_2^2
\end{array}
\end{equation}
The gradient of the validation loss with respect to the penalty parameters now follows from \eqref{implicitDifferentiation} and the chain rule. The details are given in the Appendix.

\subsubsection{Un-pooled Sparse Group Lasso}\label{sec:sgl}

The sparse group lasso is a linear regression problem that combines the $\|\cdot\|_2$ and $\|\cdot\|_1$ penalties \citep{simon2013sparse}. This method is well-suited for problems where features have a natural grouping, and only a few of the features from a few of the groups are thought to have an effect on response (e.g. genes in gene pathways). Here we consider a generalized version of sparse group lasso by considering individual penalty parameters for each group lasso penalty. This additional flexibility allows setting covariate and covariate group effects to zero by different thresholds. Hence ``un-pooled" sparse group lasso may be better at modeling covariate groups with very different distributions.

The problem setup is as follows. Given $M$ covariate groups, suppose $\boldsymbol{X}$ and $\boldsymbol \theta$ are partitioned into $\boldsymbol{X}^{(m)}$ and $\boldsymbol \theta^{(m)}$ for groups $m = 1, ... , M$. We are interested in finding the optimal regularization parameters $\boldsymbol{\lambda} = (\lambda_0, \lambda_1, ...,  \lambda_M)^\top$. The joint optimization problem is formulated as follows.
\begin{equation}
\begin{array}{c}
\min_{\boldsymbol{\lambda} \in \mathbb{R}^{M+1}_{+}} \frac{1}{2}
\left \| \boldsymbol{y}_V - \boldsymbol{X}_V \hat{\boldsymbol{\theta}}(\boldsymbol{\lambda}) \right \|^2_2 \\
\text{s.t. }
\hat{\boldsymbol{\theta}}(\boldsymbol{\lambda}) =
\argmin_{\boldsymbol{\theta}} \frac{1}{2} 
\left \| \boldsymbol{y}_T - \boldsymbol{X}_T \boldsymbol{\theta} \right \|^2_2
+ \lambda_0 \| \boldsymbol\theta \|_1
+ \sum_{m=1}^M  \lambda_m \| \boldsymbol\theta^{(m)} \|_2
+ \frac{1}{2} \epsilon \| \boldsymbol\theta \|_2^2
\end{array}
\label{eq:unpooled_sgl}
\end{equation}

By the same logic as before, the differentiable space $S_{\boldsymbol \lambda}$ is $span(\boldsymbol I_{I(\boldsymbol\lambda)})$ where $I(\boldsymbol\lambda)$ are the nonzero indices of $\hat{\boldsymbol{\theta}}(\boldsymbol{\lambda})$. Therefore the locally equivalent joint optimization problem is
\begin{equation}
\begin{array}{c}
\min_{\boldsymbol{\lambda} \in \mathbb{R}^{M+1}_{+}} \frac{1}{2} \left \| \boldsymbol{y}_V - \boldsymbol X_{V,I(\boldsymbol\lambda)} \hat{\boldsymbol\beta}(\boldsymbol{\lambda}) \right \|^2_2 \\
\text{s.t. }
\hat{\boldsymbol{\beta}}(\boldsymbol{\lambda}) = \argmin_{\boldsymbol \beta}
\frac{1}{2} \left \| \boldsymbol{y}_T - \boldsymbol{X}_{T, I(\boldsymbol\lambda)} \boldsymbol \beta \right \|^2_2
+ \lambda_0 \| \boldsymbol \beta \|_1
+ \sum_{m=1}^M \lambda_m \| \boldsymbol \beta^{(m)} \|_2
+ \frac{1}{2}\epsilon \| \boldsymbol \beta \|_2^2
\end{array}
\end{equation}
where we use the same notational shorthand $\boldsymbol{X}_{T, I(\boldsymbol\lambda)}$ and $\boldsymbol{X}_{V, I(\boldsymbol\lambda)}$ from Section \ref{sec:enet}. It is now straightforward to derive the gradient of the validation loss using \eqref{implicitDifferentiation} and the chain rule. See the appendix for details.

\subsubsection{Low-Rank Matrix Completion}\label{sec:matrix_completion}

In this example, we move away from the simple regression framework and consider matrix-valued data with partially observed entries. Our goal is to reconstruct the rest of the matrix based on the observed entries. There are many applications where such problems come up, including collaborative filtering \citep{netflix} and robust principle components analysis \citep{candes2011robust}. 
In these problems it is popular to assume a low rank structure and reconstruct the matrix by minimizing a penalized loss with a nuclear norm penalty $\|\cdot \|_*$ \citep{fazel2002matrix, srebro2004learning}. The nuclear norm -- the sum of the singular values of a matrix -- is the matrix-variate analog to the lasso. It is a non-smooth function and employing it as a penalty results in estimates that are low rank.

One extension of the matrix completion problem is to incorporate additional information about the rows and columns \citep{fithian2013scalable}. Suppose we have covariates corresponding to each row and column. Each entry of the matrix can be modeled as the sum of a low rank effect $\boldsymbol{\Gamma}$ and a linear function of the corresponding column and row covariates. Furthermore, suppose that there is a natural grouping of the row and column features. So we penalize $\boldsymbol{\Gamma}$ with the nuclear norm penalty and the linear model with group lasso penalties.

More specifically, consider an outcome matrix $\boldsymbol{M} \in \mathbb{R}^{N\times N}$. Let $\boldsymbol{X}\in \mathbb{R}^{N \times p}$ be the feature vectors for the rows and $\boldsymbol{Z} \in \mathbb{R}^{N \times p}$ be the feature vectors for columns. We assume $\boldsymbol{M}$ is composed of entries
\begin{equation}
M_{ij} = x_i \boldsymbol{\alpha} + z_j \boldsymbol{\beta} + \Gamma_{ij} + \epsilon_{ij}
\end{equation}
where $\epsilon$ are independent with mean zero. We only observe some subset of the positions in $\boldsymbol{M}$. Let the observed matrix positions be partitioned into a training set $T$, and a validation set $V$. Let $\| \cdot \|^2_T$ (resp. $\| \cdot \|^2_V$) denote the Frobenius norm over the entries in $T$ (resp. $V$). For simplicity, suppose the number of row and column feature groups, $G$, is the same, though in practice they often differ. We partition the row and column coefficient vectors $\boldsymbol{\alpha}$ and $\boldsymbol{\beta}$ into $\boldsymbol{\alpha}^{(g)}$ and $\boldsymbol{\beta}^{(g)}$ for groups $g=1,...,G$. 

We are interested in finding the optimal regularization parameters $\boldsymbol{\lambda} = (\lambda_0, \lambda_1, ...,  \lambda_{2G})^\top$. The joint optimization problem is formulated as follows
\begin{equation}
	\begin{array}{c}
		\min_{\boldsymbol{\lambda} \in \mathbb{R}^{2G+1}_{+}} \frac{1}{2}
		\left \| 
		\boldsymbol{M} 
		- \boldsymbol{X} \hat{\boldsymbol{\alpha}}(\boldsymbol{\lambda})  \boldsymbol{1}^\top 
		- (\boldsymbol{Z} \hat{\boldsymbol{\beta}}(\boldsymbol{\lambda})  \boldsymbol{1}^\top )^\top
		- \hat{\boldsymbol{\Gamma}}(\boldsymbol{\lambda})
		\right \|^2_V \\
		\text{s.t. }
		\hat{\boldsymbol{\alpha}}(\boldsymbol{\lambda}),
		\hat{\boldsymbol{\beta}}(\boldsymbol{\lambda}),
		\hat{\boldsymbol{\Gamma}}(\boldsymbol{\lambda})
		 =
		\argmin_{\boldsymbol{\alpha}, \boldsymbol{\beta}, \boldsymbol{\Gamma}} 
		\frac{1}{2} 
		\left \| 
		\boldsymbol{M} 
		- \boldsymbol{X} \boldsymbol{\alpha} \boldsymbol{1}^\top 
		- (\boldsymbol{Z} \boldsymbol{\beta} \boldsymbol{1}^\top )^\top
		- \boldsymbol{\Gamma}
		\right \|^2_T \\
		+ \lambda_0 \| \boldsymbol\Gamma \|_*
		+ \sum_{g=1}^G \lambda_g \| \boldsymbol\alpha^{(g)} \|_2
		+ \sum_{g=1}^G  \lambda_{g+G} \| \boldsymbol\beta^{(g)} \|_2
		+ \frac{\epsilon}{2}  \left (
		\| \boldsymbol\alpha \|_2^2 + \| \boldsymbol\beta \|_2^2
		+ \| \boldsymbol{\Gamma}\|^2_F
		\right )
	\end{array}
	\label{eq:matrix_comp_groups}
\end{equation}

The differentiable space of the training criterion is the product space of the differentiable spaces of the training criterion with respect to $\boldsymbol{\alpha}$, $\boldsymbol{\beta}$, and $\boldsymbol{\Gamma}$. The differentiable space with respect to $\boldsymbol{\alpha}$ is  $span(\boldsymbol{I}_{\boldsymbol{I}_r(\boldsymbol\lambda)})$ where
$$
\boldsymbol{I}_r(\boldsymbol\lambda)
= 
span(\{
i | \boldsymbol{\alpha}^{(g)} \text{ that contains index } i \text{ satisfies }
\hat{\boldsymbol{\alpha}}^{(g)}(\boldsymbol{\lambda}) \ne \boldsymbol{0}, i=1,...,p
\})
$$ 
The differentiable space with respect to $\boldsymbol{\beta}$ is $span(\boldsymbol{I}_{\boldsymbol{I}_c(\boldsymbol\lambda)})$, where $\boldsymbol{I}_c(\boldsymbol\lambda)$ is defined similarly. We show in the Appendix that the differentiable space with respect to $\boldsymbol{\Gamma}$, denoted $S_{\boldsymbol \lambda, \boldsymbol{\Gamma}}$, can be written as the span of an orthonormal basis $\{\boldsymbol{B}^{(i)}_{\boldsymbol{\lambda}}\}_{i=1}^{B}$.
The differentiable space for this problem is the product space of these three differentiable spaces
\begin{equation}
S_{\boldsymbol \lambda} 
= 
S_{\boldsymbol \lambda, \boldsymbol{\Gamma}} 
\oplus span(\boldsymbol{I}_{\boldsymbol{I}_r(\boldsymbol\lambda)})
\oplus span(\boldsymbol{I}_{\boldsymbol{I}_c(\boldsymbol\lambda)})
\label{eq:matrix_completion_diff_space}
\end{equation}
Let $J_\alpha(\boldsymbol{\lambda}) = \{g | \hat{\boldsymbol{\alpha}}^{(g)}(\boldsymbol{\lambda}) \ne \boldsymbol{0}, g = 1,...,G \}$ and define $J_\beta(\boldsymbol{\lambda})$ similarly for $\hat{\boldsymbol{\beta}}(\boldsymbol{\lambda})$.
A locally equivalent joint optimization problem is then
\begin{equation}
\begin{array}{c}
\min_{\boldsymbol{\lambda} \in \mathbb{R}^{2G+1}_{+}} \frac{1}{2}
\left \| 
\boldsymbol{M} 
- \boldsymbol{X}_{I_r(\boldsymbol{\lambda})} \hat{\boldsymbol{\eta}}(\boldsymbol{\lambda})  \boldsymbol{1}^\top 
- (\boldsymbol{Z}_{I_c(\boldsymbol{\lambda})} \hat{\boldsymbol{\gamma}}(\boldsymbol{\lambda})  \boldsymbol{1}^\top )^\top
- \sum_{i=1}^{B} \hat{b}_i(\boldsymbol{\lambda}) \boldsymbol{B}^{(i)}_{\boldsymbol{\lambda}}
\right \|^2_V \\
\text{s.t. }
\hat{\boldsymbol{\eta}}(\boldsymbol{\lambda}),
\hat{\boldsymbol{\gamma}}(\boldsymbol{\lambda}),
\hat{\boldsymbol{b}}(\boldsymbol{\lambda})
=
\argmin_{
	\boldsymbol{\eta}, \boldsymbol{\gamma}, \boldsymbol{b}
} 
\frac{1}{2} 
\left \| 
\boldsymbol{M} 
- \boldsymbol{X}_{I_r(\boldsymbol{\lambda})} \boldsymbol{\eta} \boldsymbol{1}^\top 
- (\boldsymbol{Z}_{I_c(\boldsymbol{\lambda})} \boldsymbol{\gamma} \boldsymbol{1}^\top )^\top
- \sum_{i=1}^{B} b_i \boldsymbol{B}^{(i)}_{\boldsymbol{\lambda}}
\right \|^2_T \\
+ \lambda_0  \left \| \sum_{i=1}^B b_i \boldsymbol{B}^{(i)}_{\boldsymbol{\lambda}} \right  \|_*
+ \sum_{g \in J_\alpha(\boldsymbol{\lambda})} \lambda_g \| \boldsymbol\eta^{(g)} \|_2
+ \sum_{g \in J_\beta(\boldsymbol{\lambda})} \lambda_{G+g} \| \boldsymbol\gamma^{(g)} \|_2
+ \frac{\epsilon}{2}  \left (
\| \boldsymbol\eta \|_2^2 + \| \boldsymbol\gamma \|_2^2 
+ \left  \|\sum_{i=1}^{B} b_i \boldsymbol{B}^{(i)}_{\boldsymbol{\lambda}} \right \|^2_F
\right )
\end{array}
\label{eq:matrix_comp_groups_smooth}
\end{equation}
where we use the same notational shorthand $\boldsymbol{X}_{T, I(\boldsymbol\lambda)}$ and $\boldsymbol{X}_{V, I(\boldsymbol\lambda)}$ from Section \ref{sec:enet}.

To calculate the gradient of the validation loss, we slightly modified Algorithm \ref{alg:gradDescent}. The details are given in the Appendix.

\section{Simulation Studies}\label{sec:results}

We now compare our gradient descent algorithm to gradient-free methods through simulation studies. Each simulation corresponds to a joint optimization problem given in Section \ref{sec:examples}. We tune the regularization parameters over a training/validation split using gradient descent, Nelder-Mead, and the Bayesian optimization solver from \citet{snoek2012practical} called Spearmint. For baseline comparison, we also solve a two-parameter version of the joint optimization problem using gradient descent, Nelder-Mead, Spearmint, and grid search. Each simulation was run thirty times. If the joint optimization problem had more than forty penalty parameters, we only used gradient descent to tune the parameters. Gradient-free methods are generally not recommended for tuning more than thirty parameters; Spearmint's implementation is limited to no more than forty hyperparameters and Nelder-Mead performed poorly. For details on gradient descent settings, refer to the Appendix.

We compare the efficiency of the methods by the number of times the methods solved the inner training criterion (labeled ``\# Solves" in the tables below). We set this number to 100 for all three gradient-free methods. This number is variable for gradient descent since it depends on how fast the algorithm converges.

There are two computational concerns when tuning regularization parameters by gradient descent. First, the gradient calculation can be slow for high-dimensional problems if the matrix in \eqref{eq:gradient_hessian} is large. \citet{bengio2000gradient} and \citet{foo2008efficient} suggest using a Cholesky decomposition or conjugate gradients to speed this up. However, this is not a problem for the non-smooth regression problems that we consider. The computational time to calculate the gradient of the validation loss does not grow with the number of model parameters; instead it grows with the dimension of the differentiable/local optimality space. Therefore we can efficiently calculate the gradient of the validation loss as long as the dimension of the differentiable space is small. The second concern is that the inner optimization problem must be solved to a high accuracy in order to calculate the gradient. (Recall that the gradient is derived via implicit differentiation.) To address this, we allow more iterations for solving the inner optimization problem. For a faster implementation of gradient descent, one can use a more specialized solver.

In some of the examples, the models with many penalty parameters fit by gradient descent have much smaller validation errors compared to the test errors. The difference is particularly pronounced in the un-pooled sparse group lasso example in Section~\ref{sec:simulation_sgl}. There are two reasons for this behavior. First, the additional tuning parameters increase the model space and thus the ``degrees of freedom.'' Degrees of freedom relate directly to over-optimism \citep{tibshirani2015degrees}. Traditionally one thinks of over-optimism as the difference between a model's performance on future data and its training error. In the case of hyper-parameter tuning, performance on the validation data is the analog of the ``training error.''
The second reason is that gradient descent can effectively find a near minimizer on the validation data in contrast to Nelder-mead and Spearmint. By failing to minimize the validation data, the latter two methods are similar to ending gradient descent before it has reached convergence. This technique called ``early stopping'' is another form of regularization that can control the degree of over-optimism \citep{yao2007early}. Hence the difference between the validation and test error is greater in gradient descent compared to Nelder-mead and Spearmint.

\subsection{Elastic Net}
Each dataset consists of 80 training and 20 validation observations with 250 predictors. The $\boldsymbol x_i$ were marginally distributed $N(\boldsymbol 0,\boldsymbol I)$ with $cor(x_{ij},x_{ik}) = 0.5^{|j-k|}$.
The response vector $\boldsymbol y$ was generated by
\begin{equation}
\boldsymbol y = \boldsymbol X \boldsymbol \beta + \sigma \boldsymbol \epsilon \; \text{where} \; \boldsymbol \beta = (\underbrace{1, ..., 1}_\text{size 15}, \underbrace{0, ..., 0}_\text{size 235})
\end{equation}
and $\boldsymbol \epsilon \sim N(\boldsymbol 0, \boldsymbol I)$. $\sigma$ was chosen such that the signal to noise ratio is 2. 

Grid search was performed over a $10 \times 10$ log-spaced grid from 1$e$-5 to 100. Nelder-mead and gradient descent were initialized at (0.01, 0.01) and (10, 10). Nelder-mead was allowed fifty iterations starting from each initialization point.

As shown in Table \ref{tab:elasticnet}, all the methods achieve similar validation errors. For this simple problem, the benefit for using gradient descent is not significant.

\begin{table}
\caption {\label{tab:elasticnet} Comparison of solvers for Elastic Net. Standard errors are given in parentheses.}
\centering
\begin{tabular}{| l | l | l | l | }
\hline
& Validation Error  & Test Error & \# Solves\\
\hline
Gradient Descent & 4.92 (0.41) & 5.44 (0.20) & 32.40 \\
\hline
Nelder-Mead & 4.87 (0.40) & 5.50 (0.19) & 100 \\
\hline
Spearmint & 5.15 (0.41) & 5.76 (0.22) & 100 \\
\hline
Grid Search & 5.15 (0.44) & 5.47 (0.19) & 100 \\
\hline
\end{tabular}
\end{table}

\subsection{Additive model with Smoothness and Sparsity Penalty}
\label{sec:simulation_sparse_add}
Each dataset consists of 100 training, 50 validation, and 50 test observations with $p=23$ covariates. The covariates $\boldsymbol{x_i} \in \mathbb{R}^{n}$ for $i=1,...p$ are equally spaced from -5 to 5 with a random displacement $ \delta \sim U(0, \frac{1}{300}) $ at the start and then shuffled randomly. The true model is the sum of three nonzero functions and 20 zero functions. Response $y$ was generated as follows:
\begin{equation}
\boldsymbol y = \sum\limits_{i=1}^p \boldsymbol f_i(\boldsymbol x_i) + \sigma \boldsymbol \epsilon
\label{eq:simulation_sparse_add}
\end{equation}
where $f_1(x) = 9 \sin(3x)$,
$f_2(x) = x$, 
$f_3(x) = 6 \cos(1.25 x) + 6 \sin(0.5 x + 0.5)$, and $f_i(x) \equiv 0$ for $i = 4,...,p$. $\boldsymbol \epsilon$ were drawn independently from the standard normal distribution and $\sigma$ was chosen such that the signal to noise ratio was 2.

Nelder-Mead and gradient descent were both initialized at $(10, 1, ..., 1)$ and $(0.1, 0.01, ..., 0.01)$. Nelder-Mead was allowed fifty iterations starting from each initialization point.

For baseline comparison, we also solved a two-parameter version of this joint optimization problem by pooling $\{\lambda_i\}_{i=1:p}$ into a single $\lambda_1$. Grid search was then performed over a $10 \times 10$ log-spaced grid from 1$e$-4 to $100$. 

As shown in Table \ref{tab:additive}, gradient descent returned models with the lowest test error on average. We can see that the joint optimization problem with 24 penalty parameters is sensitive to the penalty parameters chosen; Spearmint returned a result that is not only worse than gradient descent but even worse than grid search over the pooled two-parameter joint optimization problem.

\begin{figure}[ht]
\caption{Example model fits given by gradient descent for $f_1, f_2, f_3,$ and $ f_4$. The dashed lines are the true functions and the solid lines are the estimated functions. (Top left: $f_1$, top right: $f_2$, bottom left: $f_3$, bottom right: $f_4$)}
\centering
\includegraphics[height=50mm]{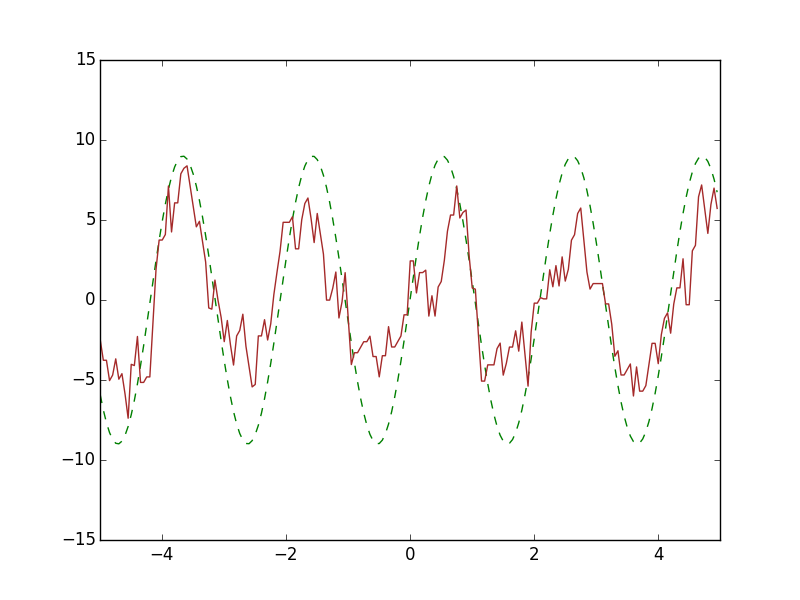}
\includegraphics[height=50mm]{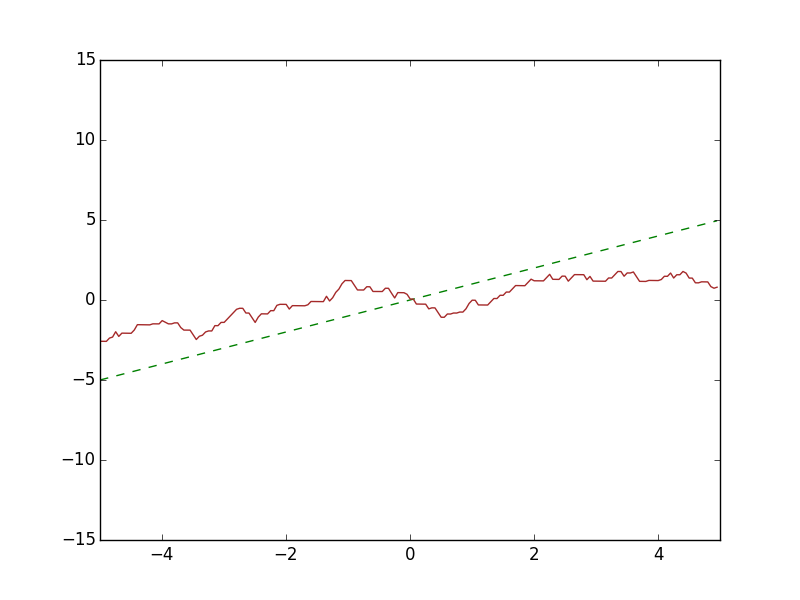}
\includegraphics[height=50mm]{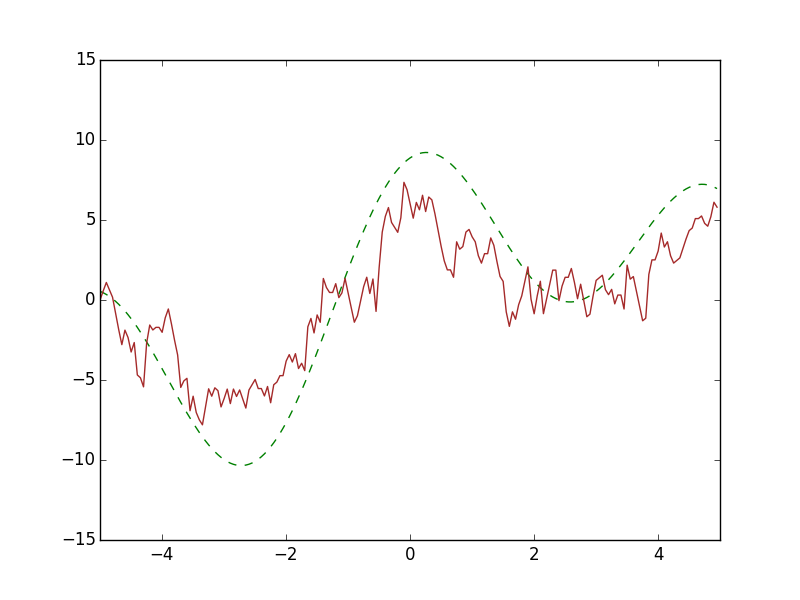}
\includegraphics[height=50mm]{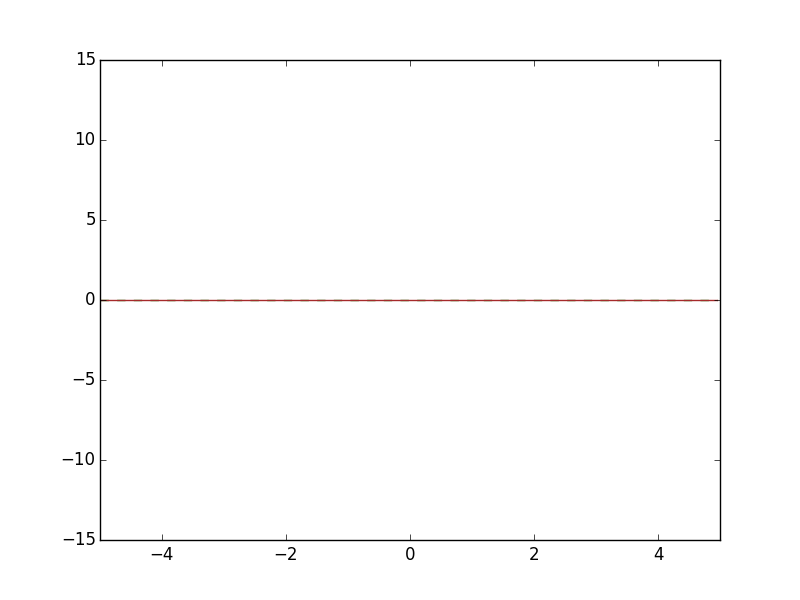}
\label{fig:additive}
\end{figure}

Table \ref{tab:additive_average_lambda} provides the average penalty parameter values for $\lambda_0, ..., \lambda_4$. Gradient descent was indeed able to determine the smoothness of the functions. It tended to choose $\lambda_1$ as the smallest and $\lambda_2$ as the largest, which corresponds to $f_1$ having the most variable slope and $f_2$ having a constant slope. In contrast, Nelder-Mead wasn't able to determine the difference in smoothness between the functions and kept all penalty parameters the same. Spearmint chose a very different set of parameters that don't seem to be appropriate for the problem. Figure \ref{fig:additive} provides example model fits given by gradient descent. The smoothness of the function estimates reflect the magnitude of the regularization parameters given in Table \ref{tab:additive}. In addition, gradient descent was able to learn that $f_4$ was the zero function.

\begin{table}
\caption {\label{tab:additive} A comparison of additive models with separate smoothness penalty parameters tuned by gradient descent, Nelder-Mead, and Spearmint vs. additive models with two penalty parameters tuned by grid search. Standard errors are given in parentheses.}
\centering
\begin{tabular}{| l | l | l | l | l | }
\hline
& \# $\lambda$ & Validation Error & Test Error & \# Solves\\
\hline
Gradient Descent & 24 & 23.87 (0.97) & 26.10 (0.86) & 13.07 \\
\hline
Nelder-Mead & 24 & 27.94 (0.90) & 28.73 (0.90) & 100 \\
\hline
Spearmint & 24 & 27.91 (1.80) & 34.02 (1.59) & 100 \\
\hline
Grid Search & 2 & 28.71 (0.97) & 29.42 (0.96) & 100 \\
\hline
\end{tabular}
\end{table}

\begin{table}
\caption {\label{tab:additive_average_lambda} Average $\lambda_i$ values for the additive models}
\centering
\begin{tabular}{| l | l | l | l | l | l | }
\hline
& $\lambda_0$ & $\lambda_1$ & $\lambda_2$ & $\lambda_3$ & $\lambda_4$\\
\hline
Gradient Descent & 9.95 & 0.40 & 1.01 & 0.86 & 1.04 \\
\hline
Nelder-Mead & 9.76 & 0.90 & 1.00 & 1.00 & 1.01 \\
\hline
Spearmint & 0.026 & 0.003 & 0.02 & 0.006 & 0.018\\
\hline
Grid Search & 4.04 & 0.38 & -- & -- & -- \\
\hline
\end{tabular}
\end{table}

\subsection{Un-pooled Sparse group lasso}\label{sec:simulation_sgl}
We ran three experiments with different numbers of covariate groups $M$ and total covariates $p$, as given in Table \ref{table:unpooled}. For each experiment, the dataset consisted of $n$ training, $n/3$ validation, and 200 test observations. The predictors $\boldsymbol X$ were generated from a standard normal distribution. The response $\boldsymbol y$ was generated by
\begin{equation}
\boldsymbol y = \sum\limits_{j=1}^3 \boldsymbol X^{(j)} \boldsymbol \beta^{(j)} + \sigma \boldsymbol \epsilon \; \text{where} \; \boldsymbol \beta^{(j)} = (1, 2, 3, 4, 5, 0, ..., 0)
\end{equation}
where $\boldsymbol \epsilon \sim N(\boldsymbol 0, \boldsymbol I)$. $\sigma$ was chosen such that the signal to noise ratio was 2. 

We compare gradient descent against all three gradient-free methods in the first experiment with 31 regularization parameters and only compare against grid search for the latter experiments with 61 and 101 regularization parameters. For all experiments, grid search solved a two-parameter version where $\{\lambda_i\}_{i=1:M}$ are pooled into a single $\lambda_1$.

For all three experiments, grid search was performed over a $10 \times 10$ log-spaced grid from 1$e$-3 to $10$. Gradient descent was initialized at $0.1 \times \boldsymbol 1$ and $\boldsymbol 1$. Nelder-Mead was initialized at the same points.

As shown in Table~\ref{table:unpooled}, the model tuned by gradient descent produced the lowest test error in all three experiments. Nelder-Mead struggled to minimize the validation error and its test error was similar to that from grid search for the two penalty-parameter problem. Interestingly, Spearmint found models with small validation error but had the highest test error.

\begin{table}
\caption{\label{table:unpooled} Un-pooled sparse group lasso and sparse group lasso tuned by gradient descent and grid search, respectively. Standard errors are given in parentheses. We abbreviated the methods as follows: Gradient Descent = GD, Nelder-Mead = NM, Spearmint = SP, Grid Search = GS}
\centering
\begin{tabular}{| l | l | l | l | l | l | }
\hline
\multicolumn{5}{|c|}{n=90, p=600, M=30}\\
\hline
& \# $\lambda$ & Validation Err & Test Err & \# Solves \\
\hline
GD & 31 & 18.88 (0.78) & 38.92 (1.47) & 40.43 (0.64) \\
\hline
NM & 31  & 47.78 (2.25) & 49.45 (1.43) & 100\\
\hline
SP & 31 &  25.91 (1.44) & 53.18 (1.54) & 100\\
\hline
GS & 2 & 47.23 (2.26) & 50.01 (1.40) & 100 \\
\hline
\multicolumn{5}{|c|}{n=90, p=900, M=60}\\
\hline
& \# $\lambda$ & Validation Error & Test Error & \# Solves \\
\hline
GD & 61 & 18.78 (1.25) & 41.88 (1.51) & 38.13 (1.48)\\
\hline
GS & 2 &  45.70 (2.27) & 51.34 (1.86) & 100 \\
\hline
\multicolumn{5}{|c|}{n=90, p=1200, M=100}\\
\hline
& \# $\lambda$ & Validation Error & Test Error & \# Solves \\
\hline
GD & 101 & 17.91 (1.44) & 47.47 (2.00) & 37.83 (1.33) \\
\hline
GS & 2 & 50.00 (2.16) & 57.14 (2.18) & 100 \\
\hline
\end{tabular}
\end{table}

\subsection{Low-Rank Matrix Completion}\label{sec:simulation_matrix}
For this experiment, we considered a $60 \times 60$ matrix. We observe two entries per row and column in the training set and one entry per row and column in the validation set. The row features were partitioned into twelve covariate groups of three covariates each, and similarly for the column features. 

The true coefficients are $\boldsymbol{\alpha}^{(g)} = g \boldsymbol{1}$ for $g = 1,...,4$ and $\boldsymbol{\beta}^{(g)} = g \boldsymbol{1}$ for $g = 1,2$. The rest of the coefficients were zero. We generated rank-one interaction matrices $\boldsymbol{\Gamma} = \boldsymbol{u}\boldsymbol{v}^\top$, where $\boldsymbol{u}$ and $\boldsymbol{v}$ were sampled from a standard normal distribution. The predictors $\boldsymbol X$ and $\boldsymbol{Z}$ were sampled from a standard normal distribution and scaled so the $l_2$ norm of $\boldsymbol{X}\boldsymbol{\alpha}\boldsymbol{1}^\top + (\boldsymbol{Z}\boldsymbol{\beta}\boldsymbol{1}^\top)^\top$ was the same as $\boldsymbol{\Gamma}$.
The noise $\boldsymbol \epsilon$ was generated from a standard normal distribution and scaled such that the signal to noise ratio was 2. 

Grid search tuned the two-parameter version of \eqref{eq:matrix_comp_groups} where $\{\lambda_i\}_{i=1:2G}$ are pooled into a single $\lambda_1$. Grid search was performed over a $10 \times 10$ log-spaced grid from 1$e$-3.5 to $-1$. Gradient descent was initialized at $0.005 \times \boldsymbol 1$ and $0.003 \times \boldsymbol 1$. Nelder-Mead was initialized at the same points.

As seen in Table \ref{table:matrix_completion}, gradient descent had the lowest average validation and test error.

\begin{table}
	\caption{\label{table:matrix_completion} Matrix Completion. Standard errors are given in parentheses. We abbreviated the methods as follows: Gradient Descent = GD, Nelder-Mead = NM, Spearmint = SP, Grid Search = GS}
	\centering
		\begin{tabular}{| l | l | l | l | l | l | }
		\hline
		& \# $\lambda$ & Validation Err & Test Err & \# Solves\\
		\hline
		GD & 25 & 0.63 (0.04) & 0.67 (0.04) & 13.00 (0.86)\\
		\hline
		NM & 25 & 0.76 (0.04)) & 0.74 (0.04) & 100 \\
		\hline
		SP & 25 & 0.65 (0.03) & 0.74 (0.04) & 100\\
		\hline
		GS & 2 & 0.71 (0.04) & 0.72 (0.04) & 100\\
		\hline
	\end{tabular}
\end{table}

\section{Application to Biological Data}\label{realDataResults}
Finally, we applied our algorithm in a real data example. More specifically, we considered the problem of finding predictive genes from gene pathways for Crohn's Disease and Ulcerative Colitis. \citet{simon2013sparse} addressed this problem using the sparse group lasso; we now compare this against applying the un-pooled sparse group lasso, where the regularization parameters were tuned using gradient descent. Since this is a classification task, the joint optimization problem is the same as \eqref{eq:unpooled_sgl} but with the logistic loss:
\begin{equation}
L\left ( \boldsymbol{y}, f_{\boldsymbol \beta(\boldsymbol\lambda)}(\boldsymbol{X}) \right ) = \sum_{i=1}^{n} y_{i} \log \left ( \frac{1}{1+\exp(-\boldsymbol x_{i}^\top \boldsymbol \beta)} \right ) + (1- y_i)\log \left (1 - \frac{1}{1+\exp(-\boldsymbol x_i^\top \boldsymbol \beta)} \right)
\end{equation}

Our dataset is from a colitis study of 127 total patients, 85 with colitis and 42 healthy controls \citep{burczynski2006molecular}. Expression data was measured for 22,283 genes on affymetrix U133A microarrays. We grouped the genes according to the 326 C1 positional gene sets from MSigDb v5.0 \citep{subramanian2005gene} and discarded 2358 genes not found in the gene set.

We randomly shuffled the data and used the first 50 observations for the training set and the remaining 77 for the test set. Five-fold cross validation was used to fit models. To tune the penalty parameters in un-pooled sparse group lasso, we initialized gradient descent at $0.5 \times \boldsymbol 1$. For sparse group lasso, we tuned the penalty parameters over a $5 \times 5$ grid 1$e$-4 to 5.

Table \ref{colitis} presents the average results from ten runs. Un-pooled sparse group lasso achieved a significantly higher classification rate and lower false negative rate compared to the sparse group lasso. The false positive rates of the two methods were not significantly different. Interestingly, un-pooled sparse group lasso found solutions that were significantly more sparse than sparse group lasso; on average, un-pooled sparse group lasso identified 9 genesets whereas sparse group lasso identified 38. These results suggest that un-pooling the penalty parameters in sparse group lasso could potentially improve interpretability.
\begin{table}
\caption{\label{colitis} Predictive genes and genesets of Ulcerative Colitis found by un-pooled sparse group lasso vs. sparse group lasso. Standard errors given in parenthesis. Gradient Descent = GD, Grid Search = GS, FP = False positive, FN = False negative}
\centering
\begin{tabular}{| l |l |l | l | l | l | l | }
\hline
& \# $\lambda$ & \% Correct  & \% FP & \% FN & \# Genesets & \# Genes  \\
\hline
GD & 327 & 89.48 (0.93) & 3.72 (1.12) & 19.90 (2.30) &  12.10 (0.98) & 55.60 (6.92) \\
\hline
GS & 2 & 86.88 (1.65) & 3.02 (0.57) & 25.34 (2.94)  & 30.80 (5.18) & 215.40 (20.69) \\
\hline
\end{tabular}
\end{table}

\section{Discussion}
In this paper we showed how to calculate the exact gradient for joint optimization problems with non-smooth penalty functions. In addition we provide an algorithm for tuning the regularization parameters using a variation of gradient descent. 

The simulation studies show that for certain problems, separating the penalty parameters can improve model performance. However, it is crucial that the regularization parameters be tuned appropriately. For the same joint optimization problem with many penalty parameters, gradient descent is able to return good model fits whereas gradient-free methods like Nelder-Mead and Bayesian optimization tend to fail. In fact, when the optimization method is unable to tune the regularization parameters appropriately, we find that a simple grid search over the pooled two-parameter joint optimization problem can result in better models.

Through our simulation studies, we did find that this gradient-based approach depends on solving the inner optimization problem to a higher level of accuracy than needed for gradient-free approaches. More work could be done to investigate the degree of accuracy required for gradient descent to still be effective.

Since our algorithm depends on the validation loss being smooth almost everywhere, a potential concern is that the validation loss may not be differentiable at the solution of the joint optimization problem. We believe that this scenario occurs with measure zero. For a more detailed discussion, refer to the Appendix.

Finally, an open theoretical question is how much the model complexity increases when too many penalty parameters are introduced. Similar to how model parameters can overfit to their training data, it is possible for the penalty parameters to overfit to the training and validation data.

\section{Supplementary Files}
\begin{description}
	\item[Appendix:] The appendix contains a proof for Theorem~\ref{thethrm}, detailed gradient derivations for the examples in Section~\ref{sec:examples}, and additional simulation results.
	\item[Python Code:] Code used in Sections~\ref{sec:results} and \ref{realDataResults} can be downloaded from \url{https://github.com/jjfeng/nonsmooth-joint-opt}.
\end{description}

	\appendix
	
	\section{Appendix}
	
	\subsection{$K$-fold Cross Validation}
	We can perform joint optimization for $K$-fold cross validation by reformulating the problem. Let $(\boldsymbol y, \boldsymbol{X})$ be the full data set. We denote the $k$th fold as $(\boldsymbol y_{k}, \boldsymbol{X}_{k})$ and its complement as $(\boldsymbol y_{-k}, \boldsymbol{X}_{-k})$. Then the objective of this joint optimization problem is the average validation cost across all $K$ folds:
	\begin{equation}
	\begin{array}{c}
	\argmin_{\boldsymbol{\lambda} \in \Lambda} \frac{1}{K} \sum_{k=1}^K L(\boldsymbol{y}_{k}, f_{\hat{\boldsymbol \theta}^{(k)}(\boldsymbol{\lambda})}(\boldsymbol{X}_k)) \\
	\text{s.t. } {\hat{\boldsymbol \theta}^{(k)}(\boldsymbol{\lambda})} = \argmin_{\boldsymbol \theta \in \Theta} L(\boldsymbol{y}_{-k}, f_{\boldsymbol \theta} (\boldsymbol{X}_{-k})) + \sum\limits_{i=1}^J \lambda_i P_i(\boldsymbol \theta) \text{ for } k=1,...,K
	\end{array}
	\label{jointoptFullCV}
	\end{equation}
	
	\subsection{Proof of Theorem~\ref{thethrm}}
	
	\begin{proof}
		We will show that for a given $\boldsymbol \lambda_0$ that satisfies the given conditions, the validation loss is continuously differentiable within some neighborhood of $\boldsymbol \lambda_0$.  It then follows that if the theorem conditions hold true for almost every $\boldsymbol \lambda$, then the validation loss is continuously differentiable with respect to $\boldsymbol \lambda$ at almost every $\boldsymbol \lambda$.
		
		Suppose the theorem conditions are satisfied at $\boldsymbol \lambda_0$. Let $\boldsymbol B'$ be an orthonormal set of basis vectors that span the differentiable space $\Omega^{L_T}(\hat {\boldsymbol \theta}(\boldsymbol \lambda_0), \boldsymbol \lambda_0)$ with the subset of vectors $\boldsymbol B$ that span the model parameter space.
		
		Let $\tilde L_T(\boldsymbol \theta,\boldsymbol \lambda)$ be the gradient of $L_T(\cdot, \boldsymbol \lambda)$ at $\boldsymbol \theta$ with respect to the basis $\boldsymbol B$:
		\begin{equation}
		\tilde L_T(\boldsymbol \theta,\boldsymbol \lambda) = _{\boldsymbol B}\nabla L_T(\cdot, \boldsymbol \lambda) |_{\boldsymbol \theta}
		\end{equation}
		
		Since $\hat {\boldsymbol \theta}(\boldsymbol \lambda_0)$ is the minimizer of the training loss, the gradient of $L_T(\cdot, \boldsymbol \lambda_0)$ with respect to the basis $\boldsymbol B$ must be zero at $\hat {\boldsymbol \theta}(\boldsymbol \lambda_0)$:
		\begin{equation}
		_{\boldsymbol B}\nabla L_T(\cdot, \boldsymbol \lambda_0)|_{\hat {\boldsymbol \theta}(\boldsymbol \lambda_0)} = \tilde L_T(\hat {\boldsymbol \theta}(\boldsymbol \lambda_0), \boldsymbol \lambda_0) = 0
		\end{equation}
		
		From our assumptions, we know that there exists a neighborhood $W$ containing $\boldsymbol \lambda_0$ such that $\tilde L_T$ is continuously differentiable along directions in the differentiable space $\Omega^{L_T}(\hat {\boldsymbol \theta}(\boldsymbol \lambda_0), \boldsymbol \lambda_0)$. Also, the Jacobian matrix $D \tilde L_T(\cdot, \boldsymbol \lambda_0)|_{\hat {\boldsymbol \theta}(\boldsymbol \lambda_0)}$ with respect to basis $\boldsymbol B$ is nonsingular. Therefore, by the implicit function theorem, there exist open sets $U \subseteq W$ containing $\boldsymbol \lambda_0$ and $V$ containing $\hat {\boldsymbol \theta}(\boldsymbol \lambda_0)$ and a continuously differentiable function $\gamma: U \rightarrow V$ such that for every $\boldsymbol \lambda \in U$, we have that 
		\begin{equation}
		\tilde L_T(\gamma(\boldsymbol \lambda), \boldsymbol \lambda) = \nabla_{B} L_T(\cdot, \boldsymbol \lambda)|_{\gamma(\boldsymbol \lambda)} = 0
		\end{equation}
		That is, we know that $\gamma(\boldsymbol \lambda)$ is a continuously differentiable function that minimizes $L_T(\cdot, \boldsymbol \lambda)$ in the differentiable space  $\Omega^{L_T}(\hat {\boldsymbol \theta}(\boldsymbol \lambda_0), \boldsymbol \lambda_0)$.
		Since we assumed that the differentiable space is a local optimality space of $L_T(\cdot, \boldsymbol \lambda)$ in the neighborhood $W$, then for every $\boldsymbol \lambda \in U$, 
		\begin{equation}
		\hat {\boldsymbol \theta}(\boldsymbol \lambda) =
		\argmin_{\boldsymbol \theta} L_T(\boldsymbol \theta, \boldsymbol \lambda) =
		\argmin_{\boldsymbol \theta \in \Omega^{L_T}(\hat {\boldsymbol \theta}(\boldsymbol \lambda_0), \boldsymbol \lambda_0)} L_T(\boldsymbol \theta, \boldsymbol \lambda) =
		\gamma(\boldsymbol \lambda)
		\end{equation}
		Therefore, we have shown that if $\boldsymbol \lambda_0$ satisfies the assumptions given in the theorem, the fitted model parameters $\hat {\boldsymbol \theta}(\boldsymbol \lambda)$ is a continuously differentiable function within a neighborhood of $\boldsymbol \lambda_0$. We can then apply the chain rule to get the gradient of the validation loss.
	\end{proof}
	
	\subsection{Regression Examples}
	
	\subsubsection{Elastic Net}\label{enet_conditions}
	We show that the joint optimization problem for the Elastic Net satisfies all three conditions in Theorem~\ref{thethrm}:
	\begin{itemize}
		\item[] Condition 1: The elastic net solution paths are piecewise linear \citep{zou2003regression}, which means that the nonzero indices of the elastic net estimates stay locally constant for almost every $\boldsymbol{\lambda}$. Therefore, $S_{\boldsymbol{\lambda}}$ as defined in Section~\ref{sec:enet}  is a local optimality space for $L_T(\cdot, \boldsymbol{\lambda})$.
		\item[] Condition 2: We only need to establish that the $\ell_1$ penalty is twice-continuously differentiable in the directions of $S_{\boldsymbol{\lambda}}$ since the quadratic loss function and the ridge penalty are both smooth. The absolute value function is twice-continuously differentiable everywhere except at zero. Hence the training criterion is smooth when restricted to $S_{\boldsymbol{\lambda}}$.
		\item[] Condition 3: The Hessian matrix of $L_T(\cdot, \boldsymbol{\lambda})$ with respect to $\boldsymbol I_{I(\boldsymbol \lambda)}$ is $\boldsymbol I_{I(\boldsymbol \lambda)}^\top \boldsymbol{X}_{T}^\top \boldsymbol{X}_{T} \boldsymbol I_{I(\boldsymbol \lambda)} + \lambda_2 \boldsymbol{I}$. The first summand is positive semi-definite. As long as $\lambda_2 > 0$, the contribution of the identity matrix ensures the Hessian is positive definite.
	\end{itemize}
	
	\subsubsection{Additive Models with Sparsity and Smoothness Penalties}
	\label{sec_appendix:sparse_add_models}
	We use the notation in Section~\ref{sec:additive}. 
	In addition, let  $|J(\boldsymbol{\lambda})|$ be the number of elements in $J(\boldsymbol{\lambda})$. Let
	\begin{equation}
	\boldsymbol{U} = \begin{bmatrix}
	\boldsymbol {U}^{(i_1)} & ... & \boldsymbol {U}^{(i_{|J(\boldsymbol \lambda)|})}
	\end{bmatrix}
	\end{equation}
	where $i_\ell \in J(\boldsymbol \lambda)$. 
	
	The gradient of the validation loss is
	\begin{equation}
	\nabla_{\boldsymbol \lambda} L(\boldsymbol{y_V}, f_{\hat{\boldsymbol{\theta}}(\boldsymbol{\lambda})}(\boldsymbol{X_V})) =
	-
	\left (
	\boldsymbol{I}_V \sum_{i\in J(\boldsymbol\lambda)}  \boldsymbol {U}^{(i)} \frac{\partial}{\partial \boldsymbol\lambda} \hat{\boldsymbol{\beta}}^{(i)}(\boldsymbol{\lambda})
	\right )^\top
	\left (
	\boldsymbol{y}_V - \boldsymbol{I}_V \sum_{i\in J(\boldsymbol\lambda)} \boldsymbol {U}^{(i)} \hat{\boldsymbol{\beta}}^{(i)} (\boldsymbol\lambda)
	\right )
	\end{equation}
	where
	\begin{equation}
	\frac{\partial}{\partial \boldsymbol{\lambda}} 
	\hat{\boldsymbol{\beta}}(\boldsymbol{\lambda})
	= 
	\boldsymbol{H}(\boldsymbol\lambda)^{-1}
	\begin{bmatrix}
	\boldsymbol{C}_0(\hat{\boldsymbol{\beta}}(\boldsymbol{\lambda}))
	&
	\boldsymbol C \left (\hat{\boldsymbol \beta}( \boldsymbol \lambda) \right )
	\end{bmatrix}
	\label{eq:additive_gradient}
	\end{equation}
	
	The Hessian $\boldsymbol{H}(\boldsymbol\lambda)$ is
	\begin{equation}
	\boldsymbol{H}(\boldsymbol\lambda)
	= \boldsymbol{U}^\top \boldsymbol I_T^\top \boldsymbol I_T \boldsymbol{U}
	+ \lambda_0 diag  \left ( \left \{
	\frac{1}{||\boldsymbol {U}^{(i)}  \hat{\boldsymbol{\beta}}^{(i)} (\boldsymbol \lambda)||_2} \left (
	\boldsymbol I - \frac{\hat{\boldsymbol{\beta}}^{(i)} (\boldsymbol \lambda) \hat{\boldsymbol{\beta}}^{(i)\top} (\boldsymbol \lambda)}{||\boldsymbol {U}^{(i)}  \hat{\boldsymbol{\beta}}^{(i)} (\boldsymbol \lambda)||_2^2}
	\right )
	\right \}_{i \in J(\boldsymbol{\lambda})}
	\right )
	+ \epsilon \boldsymbol I
	\label{eq:add_hessian}
	\end{equation}
	The vector $\boldsymbol{C}_0(\hat{\boldsymbol{\beta}}(\boldsymbol{\lambda}))$ is a vertical stack of the vectors 
	$$
	\frac{\hat{\boldsymbol{\beta}}^{(i)}(\boldsymbol \lambda)}{\left \| \boldsymbol {U}^{(i)}  \hat{\boldsymbol{\beta}}^{(i)} (\boldsymbol \lambda)\right \|_2}
	$$
	for $i \in J(\boldsymbol{\lambda})$.
	The matrix $\boldsymbol C(\hat{\boldsymbol \beta}( \boldsymbol \lambda))$ has columns $i = 1,...,p$
	\begin{equation}
	\boldsymbol{C}_i(\hat{\boldsymbol \beta}( \boldsymbol \lambda))
	= \begin{cases}
	\begin{bmatrix}
	\boldsymbol{0} \\
	\boldsymbol {U}^{(i)\top}  \boldsymbol{D}^{(2)\top}_{\boldsymbol{x}_i} 
	sgn \left ( \boldsymbol{D}^{(2)}_{\boldsymbol{x}_i} \boldsymbol {U}^{(i)} \hat{\boldsymbol{\beta}}^{(i)} ( \boldsymbol \lambda) \right ) \\
	\boldsymbol{0} \\
	\end{bmatrix}
	& \text{ for } i \in J(\boldsymbol \lambda) \\
	\boldsymbol{0}
	& \text{ for } i \not\in J(\boldsymbol \lambda) \\
	\end{cases}
	\end{equation}
	
	Now we check that all three conditions are satisfied. 
	\begin{itemize}
		\item[] Condition 1: It seems likely that the space spanned by $S_{\boldsymbol{\lambda}}$ is a local optimality space, though we are unable to formally prove this. The training criterion for this problem is composed of generalized lasso penalties and a group lasso penalties. For the generalized lasso, \citet{tibshirani2011solution} proved that the solution path is smooth almost everywhere. For the group lasso, there is empirical evidence that the active set is locally constant almost everywhere with respect to the penalty parameter \citep{yuan2006model}, but this has not been formally proven. \citet{vaiter2012degrees} showed that the active set is locally constant with respect to the response; we suspect similar techniques could be used to prove our hypothesis.
		\item[] Condition 2:  We only need to establish that the generalized lasso and group lasso penalties are twice-continuously differentiable in the directions of $S_{\boldsymbol{\lambda}}$ since the rest of the training criterion is smooth. 
		$\| \boldsymbol{D} \boldsymbol{\theta} \|_1$ is not differentiable at the points where $\boldsymbol{D} \boldsymbol{\theta}$ has zero elements. We must therefore restrict the derivatives to be taken in directions such that the zero elements of $\boldsymbol{D} \boldsymbol{\theta}$ remain constant. The $\ell_2$ norm is twice-continuously differentiable everywhere except at the zero vector. Hence the training criterion is smooth when restricted to the differentiable space $S_{\boldsymbol{\lambda}}$ specified in Section~\ref{sec:additive}.
		\item[] Condition 3: The Hessian matrix in \eqref{eq:add_hessian} is the sum of positive semi-definite matrices. As long as $\epsilon > 0$, the contribution of the last summand $\epsilon \boldsymbol{I}$ will make the Hessian matrix positive-definite.
	\end{itemize}
	
	\subsubsection{Un-pooled Sparse Group Lasso}
	
	The gradient of the validation loss with respect to the penalty parameters is
	\begin{equation}
	\nabla_{\boldsymbol \lambda} L(\boldsymbol{y_V}, f_{\hat{\boldsymbol{\theta}}(\boldsymbol{\lambda})}(\boldsymbol{X_V})) =
	-\left (
	\boldsymbol{X}_{V, I(\boldsymbol\lambda)}
	\frac{\partial}{\partial \boldsymbol\lambda} \hat{\boldsymbol{\beta}}(\boldsymbol{\lambda})
	\right )^\top
	\left (
	\boldsymbol{y}_V - \boldsymbol{X}_{V, I(\boldsymbol\lambda)} \hat{\boldsymbol{\beta}}(\boldsymbol{\lambda})
	\right )
	\end{equation}
	where 
	\begin{equation}
	\frac{\partial}{\partial \boldsymbol \lambda} \hat{\boldsymbol{\beta}}(\boldsymbol{\lambda})
	=
	- \boldsymbol H(\boldsymbol \lambda)^{-1}
	\begin{bmatrix}
	\boldsymbol C(\hat{\boldsymbol \beta}(\boldsymbol \lambda)) & sgn(\hat {\boldsymbol \beta}(\boldsymbol \lambda))
	\end{bmatrix}
	\label{eq:unpooled_sgl_grad}
	\end{equation}
	The Hessian $\boldsymbol H(\boldsymbol \lambda)$ is
	\begin{equation}
	\boldsymbol{H}(\boldsymbol\lambda) =
	\frac{1}{n} \boldsymbol X_{T, I(\boldsymbol \lambda)}^\top \boldsymbol X_{T, I(\boldsymbol \lambda)}
	+ diag\left(
	\frac{\lambda_m}{|| \boldsymbol \theta^{(m)}||_2}
	\left (
	\boldsymbol I - 
	\frac{\boldsymbol \theta^{(m)} \boldsymbol \theta^{(m) \top}}{|| \boldsymbol \theta^{(m)}||_2^2}
	\right )
	\right)
	+ \epsilon \boldsymbol I
	\label{eq:sgl_hessian}
	\end{equation}
	The matrix $\boldsymbol C(\hat {\boldsymbol \beta}(\boldsymbol \lambda))$ in \eqref{eq:unpooled_sgl_grad} has columns $m=1,2...,M$ 
	\begin{equation}
	\boldsymbol{C}_i(\hat{\boldsymbol \beta}( \boldsymbol \lambda))
	=
	\begin{bmatrix}
	\boldsymbol 0\\
	\frac{\hat {\boldsymbol \beta}^{(m)}(\boldsymbol \lambda)}{||\hat{\boldsymbol \beta}^{(m)}(\boldsymbol \lambda)||_2}\\
	\boldsymbol 0\\
	\end{bmatrix}
	\end{equation}
	where $\boldsymbol 0$ are the appropriate dimensions.
	
	The logic for checking all three conditions in Theorem~\ref{thethrm} is similar to the other examples:
	\begin{itemize}
		\item[] Condition 1: We hypothesize that the differentiable space $S_{\boldsymbol{\lambda}}$ is also a local optimality space, though we have not formally proven this fact. We suspect this is true for the same reasons discussed in Section \ref{sec_appendix:sparse_add_models}.
		\item[] Condition 2: The $\ell_1$ and $\ell_2$ penalties are twice-differentiable when restricted to $S_{\boldsymbol{\lambda}}$ for the same reasons discussed in Section~\ref{sec_appendix:sparse_add_models}. 
		\item[] Condition 3: The Hessian matrix in \eqref{eq:sgl_hessian} is the sum of positive semi-definite matrices. It is positive definite for any $\epsilon > 0$ due to the last summand $\epsilon \boldsymbol{I}$. 
	\end{itemize}
	
	\subsubsection{Low-rank Matrix Completion}
	Here we derive the differentiable space of the training criterion with respect to $\boldsymbol{\Gamma}$. At $\boldsymbol{\lambda}$, suppose the fitted interaction matrix $\hat{\boldsymbol{\Gamma}}(\boldsymbol{\lambda})$ has a singular value decomposition $\hat{\boldsymbol{U}}(\boldsymbol{\lambda}) \text{diag}(\hat{\boldsymbol{\sigma}}(\boldsymbol{\lambda})) \hat{\boldsymbol{V}}^\top(\boldsymbol{\lambda})$. We denote the $i$th singular value/vector with subscript $i$. Then the differentiable  space with respect to $\boldsymbol{\Gamma}$ at $\hat{\boldsymbol{\Gamma}}(\boldsymbol{\lambda})$ is
	\begin{align}
	S_{\boldsymbol{\lambda}, \boldsymbol{\Gamma}} & = 
	\left\{
	\boldsymbol{B} \in \mathbb{R}^{N \times N}
	\middle |
	\hat{\boldsymbol{U}}_i^\top(\boldsymbol{\lambda}) \boldsymbol{B} \hat{\boldsymbol{V}}_i(\boldsymbol{\lambda}) = 0 \quad \forall i \text{ s.t. } \sigma_i  = 0
	\right\}
	\\
	& = 
	\text{span} \left (
	\left\{
	\hat{\boldsymbol{U}}_i(\boldsymbol{\lambda}) \boldsymbol{b}_u^\top + \boldsymbol{b}_v \hat{\boldsymbol{V}}_i^\top(\boldsymbol{\lambda})
	\middle |
	\boldsymbol{b}_u, \boldsymbol{b}_v \in \mathbb{R}^{N},
	\sigma_i \ne 0
	\right\}
	\right )
	\label{eq:diff_space_gamma}
	\end{align}
	The proof is a direct application of Theorem 1 in \citet{watson1992characterization}. The following lemma adapts his results for our purposes. Note that if a matrix can be written as a univariate function $\tilde{\boldsymbol{\Gamma}}(\epsilon)$, its singular values and singular vectors can be numbered such that they are each a function of $\epsilon$, e.g. $\sigma_i(\epsilon)$, $\boldsymbol{U}_i(\epsilon)$, and $\boldsymbol{V}_i(\epsilon)$ \citep{rellich1969perturbation}.
	\begin{lemma}
		Suppose $\boldsymbol{\Gamma} \in \mathbb{R}^{N\times N}$ has a singular value decomposition $\boldsymbol{U}\text{diag}(\boldsymbol{\sigma})\boldsymbol{V}$.
		Let 
		\begin{align}
		\mathcal{B} & = 
		\left\{
		\boldsymbol{B} \in \mathbb{R}^{N \times N}
		\middle |
		{\boldsymbol{U}}_i^\top \boldsymbol{B} {\boldsymbol{V}}_i = 0 \quad \forall i \text{ s.t. } \sigma_i  = 0
		\right\}
		\end{align}
		The directional derivative of the nuclear norm $\| \cdot \|_*$ at $\boldsymbol{\Gamma}$ along $\boldsymbol{B} \in \mathcal{B}$ is
		\begin{equation}
		\lim_{\epsilon\rightarrow 0^+} \frac{\| \boldsymbol{\Gamma} + \epsilon\boldsymbol{B} \|_* - \| \boldsymbol{\Gamma} \|_*}{\epsilon}
		=
		\sum_{i=1}^N \boldsymbol{U}_i^\top \boldsymbol{B} \boldsymbol{V}_i 1_{[\sigma_i \ne 0]}
		\label{eq:direct_deriv_nuclear_norm}
		\end{equation}
		
		Moreover, let the eigenvalues be numbered such that $\sigma_{i, \boldsymbol{B}}(\epsilon)$ denotes the $i$th singular value of $\boldsymbol{\Gamma} + \epsilon \boldsymbol{B}$. Then
		\begin{equation}
		\mathcal{B} = \left \{
		\boldsymbol{B} \in \mathbb{R}^{N \times N}
		\middle |
		\left . \frac{d \sigma_{i, \boldsymbol{B}} (\epsilon )}{d\epsilon} \right |_{\epsilon = 0} = 0  \quad\forall i \text{ s.t. } \sigma_i= 0
		\right \}
		\label{eq:const_singular_values}
		\end{equation}
		
	\end{lemma}
	
	Now we derive the gradient of the validation loss with respect to the penalty parameters. One approach would be to follow Algorithm~\ref{alg:gradDescent} exactly, which requires us to find an orthonormal basis of \eqref{eq:diff_space_gamma}. 
	An alternative approach is to use the result in \eqref{eq:const_singular_values}: the differentiable space is the set of directions where the zero singular values remain locally constant. Assuming Condition~\ref{condn:local_is_diff} holds, we only need to consider interaction matrices with rank at most $r = \text{rank}(\hat{\boldsymbol{\Gamma}}(\boldsymbol{\lambda}))$. Hence a locally equivalent training criterion is:
	\begin{align}
	\begin{split}
	\argmin_{
		\substack{%
			\boldsymbol{\eta}, \boldsymbol{\gamma} \\
			\boldsymbol{\Gamma} = \boldsymbol{U}diag(\boldsymbol{\sigma}) \boldsymbol{V}^\top\\
			\boldsymbol{U}, \boldsymbol{V} \in \mathbb{R}^{N\times r},
			\boldsymbol{\sigma} \in \mathbb{R}^{r}
		}
	}
	& 
	\frac{1}{2} 
	\left \| 
	\boldsymbol{M} 
	- \boldsymbol{X}_{I_r(\boldsymbol{\lambda})} \boldsymbol{\eta} \boldsymbol{1}^\top 
	- (\boldsymbol{Z}_{I_c(\boldsymbol{\lambda})} \boldsymbol{\gamma} \boldsymbol{1}^\top )^\top
	- \boldsymbol{\Gamma}
	\right \|^2_T
	+ \lambda_0  \left \| \boldsymbol{\Gamma} \right  \|_* \\
	& + \sum_{g\in J_{\boldsymbol{\alpha}}(\boldsymbol{\lambda})}  \lambda_g \| \boldsymbol\eta^{(g)} \|_2
	+ \sum_{g\in J_{\boldsymbol{\beta}}(\boldsymbol{\lambda})}  \lambda_{G+g} \| \boldsymbol\gamma^{(g)} \|_2
	+ \frac{1}{2} \epsilon \left (
	\| \boldsymbol\eta \|_2^2 + \| \boldsymbol\gamma \|_2^2 
	+ \left  \| \boldsymbol{\Gamma} \right \|^2_F
	\right )
	\label{eq:matrix_comp_groups_svd_smooth}
	\end{split}
	\\
	& 
	\text{s.t. } 
	\boldsymbol{V}^\top \boldsymbol{V} = \boldsymbol{I}
	\text{ and } \boldsymbol{U}^\top \boldsymbol{U} = \boldsymbol{I}
	\label{eq:orthonormal_constraints}
	\end{align}
	The locally equivalent training criterion is now smooth at its minimizer.
	The gradient optimality conditions with respect to $\boldsymbol{\Gamma}$ can be taken with respect to the basis 
	\begin{align}
	\left \{
	\hat{\boldsymbol{U}}_i(\boldsymbol{\lambda}) \boldsymbol{e}_j^\top  | i = 1,...,r; j = 1,...,N
	\right \}
	\cup
	\left \{
	\boldsymbol{e}_j \hat{\boldsymbol{V}}_i(\boldsymbol{\lambda})^\top | i = 1,...,r; j = 1,...,N
	\right \}
	\label{eq:mat_completion_gamma_basis}
	\end{align}
	Note that this basis is quite different from that used in Algorithm \ref{alg:gradDescent}; it is allowed to vary with $\boldsymbol{\lambda}$ and its elements are not orthonormal. The benefit of this alternative approach is that the gradient optimality condition for $\boldsymbol{\Gamma}$ is easy to derive. Taking the gradient with respect to the directions in \ref{eq:mat_completion_gamma_basis}, we get:
	\begin{align}
	\begin{split}
	\boldsymbol{0} & = 
	- \hat{\boldsymbol{U}}(\boldsymbol{\lambda})^\top
	\left (
	\boldsymbol{M} 
	- \boldsymbol{X}_{I_r(\boldsymbol{\lambda})} \hat{\boldsymbol{\eta}}(\boldsymbol{\lambda}) \boldsymbol{1}^\top 
	- (\boldsymbol{Z}_{I_c(\boldsymbol{\lambda})} \hat{\boldsymbol{\gamma}}(\boldsymbol{\lambda})  \boldsymbol{1}^\top )^\top
	- \hat{\boldsymbol{U}}(\boldsymbol{\lambda})\text{diag}(\hat{\boldsymbol{\sigma}}(\boldsymbol{\lambda})) \hat{\boldsymbol{V}}(\boldsymbol{\lambda})^\top
	\right )_T
	\\
	& \qquad + \lambda_0 \hat{\boldsymbol{V}}(\boldsymbol{\lambda})^\top
	+ \epsilon \text{diag}(\hat{\boldsymbol{\sigma}}(\boldsymbol{\lambda})) \hat{\boldsymbol{V}}(\boldsymbol{\lambda})^\top
	\end{split}
	\label{eq:grad_opt_matrix_left}
	\\
	\begin{split}
	\boldsymbol{0} & = -\left (
	\boldsymbol{M} 
	- \boldsymbol{X}_{I_r(\boldsymbol{\lambda})} \hat{\boldsymbol{\eta}}(\boldsymbol{\lambda}) \boldsymbol{1}^\top 
	- (\boldsymbol{Z}_{I_c(\boldsymbol{\lambda})} \hat{\boldsymbol{\gamma}}(\boldsymbol{\lambda})  \boldsymbol{1}^\top )^\top
	- \hat{\boldsymbol{U}}(\boldsymbol{\lambda})\text{diag}(\hat{\boldsymbol{\sigma}}(\boldsymbol{\lambda})) \hat{\boldsymbol{V}}(\boldsymbol{\lambda})^\top
	\right )_T
	\hat{\boldsymbol{V}}(\boldsymbol{\lambda})
	\\
	& \qquad + \lambda_0 \hat{\boldsymbol{U}}(\boldsymbol{\lambda})
	+ \epsilon \hat{\boldsymbol{U}}(\boldsymbol{\lambda}) \text{diag}(\hat{\boldsymbol{\sigma}}(\boldsymbol{\lambda}))
	\end{split}
	\label{eq:grad_opt_matrix_comp}
	\end{align}
	where $(\cdot)_T$ zeroes out matrix elements that are not observed in the training set. The gradient optimality conditions with respect to $\boldsymbol{\eta}$ and $\boldsymbol{\gamma}$ are derived using the usual procedure. To get the partial derivatives of the fitted values with respect to $\boldsymbol{\lambda}$, we implicitly differentiate the gradient optimality conditions, as well as \eqref{eq:orthonormal_constraints}, with respect to $\boldsymbol{\lambda}$ and solve the resulting system of linear equations. The gradient of the validation loss with respect to the penalty parameters is straightforward to calculate once the partial derivatives are obtained. However, we omit this tedious calculation.
	
	We now show that the conditions in Theorem~\ref{thethrm} are satisfied.
	\begin{itemize}
		\item[] Condition 1: We hypothesize that the differentiable space $S_{\boldsymbol{\lambda}}$ defined in \eqref{eq:matrix_completion_diff_space} is also a local optimality space $\boldsymbol{\lambda}$. For the group lasso penalties, we use the same reasons mentioned in \ref{sec_appendix:sparse_add_models} to justify this hypothesis. For the nuclear norm penalty, it has been observed empirically that small perturbations in the penalty parameter result in matrices with similar rank \citep{mazumder2010spectral}. This supports our belief that $S_{\boldsymbol{\lambda}, \boldsymbol{\Gamma}}$ is a local optimality space with respect to $\boldsymbol{\Gamma}$ at $\boldsymbol{\lambda}$.
		\item[] Condition 2: The only non-smooth components of the training criterion are the group lasso and nuclear norm penalties. The group lasso penalty is twice-differentiable when restricted to the differentiable space, using the same reasoning in Section~\ref{sec_appendix:sparse_add_models}. From \eqref{eq:direct_deriv_nuclear_norm}, we see that the nuclear norm $\|\boldsymbol{\Gamma}\|_{*}$ is also twice-differentiable with respect to $\boldsymbol{\Gamma}$ when restricted to $S_{\boldsymbol{\lambda}, \boldsymbol{\Gamma}}$.
		
		\item[] Condition 3: The differentiable space for the training criterion with respect to $\boldsymbol{\Gamma}$ is a linear space. Therefore there exists some orthonormal basis of the differentiable space. Since the training criterion is the sum of convex functions with ridge penalties on all the variables, the Hessian of the training criterion is positive definite for any $\epsilon > 0$.
		
	\end{itemize}
	
	\subsection{Gradient Descent Details}\label{sec:alg_details}
	Here we discuss our choice of step size and convergence criterion in gradient descent.
	
	There are many possible choices for our step size sequence $\{t^{(k)}\}$ \citep{boyd2004convex}. We chose a backtracking line, which we describe here briefly. Let the criterion function be $L:\mathbb{R}^n \rightarrow \mathbb{R}$. Suppose that the descent algorithm is currently at point $x$ with descent direction $\Delta x$. The algorithm is given below. It depends on constants $\alpha  \in (0, 0.5)$ and $\beta \in (0,1)$.
	\begin{algorithm}
		\caption{Backtracking Line Search} \label{alg:backtracking}
		\begin{algorithmic}
			\STATE{Initialize $t= 1$.} \\
			\WHILE{$L(\boldsymbol x + t \boldsymbol \Delta \boldsymbol x) > L(\boldsymbol x) + \alpha t \nabla L(\boldsymbol x)^T \boldsymbol \Delta \boldsymbol x$}
			\STATE{Update $t := \beta t$}
			\ENDWHILE
		\end{algorithmic}
	\end{algorithm}
	In our examples initial step size was 1, and we backtrack with parameters $\alpha = 0.001$ and $\beta = 0.1$. During gradient descent, it is possible that the step size will result in a negative regularization parameter; we reject any step that would set a regularization parameter to below a minimum threshold of $1e$-6.
	
	Our convergence criterion is based on the change in our validation loss between iterates. More specifically, we stop our algorithm when
	\[
	L \left( \boldsymbol{y}_V, f_{\hat{\boldsymbol \theta}(\boldsymbol{\lambda}^{(k+1)})}(\boldsymbol{X}_V)\right) -
	L \left( \boldsymbol{y}_V, f_{\hat{\boldsymbol \theta}(\boldsymbol{\lambda}^{(k)})}(\boldsymbol{X}_V)\right) \leq \delta
	\]
	For the results in this manuscript we use $\delta = 0.0005$.
	
	\subsection{Sensitivity to initialization points}
	Since the results of gradient descent and Nelder-Mead depend on their initialization points, we ran a simulation to see how sensitive the methods were to where they were initialized and how many initializations were used.
	
	We tested a smaller version of the joint optimization problem in  Section~\ref{sec:additive}. Here we use 60 training, 30 validation, and 30 test observations and $p = 15$ covariates. The response was generated from \eqref{eq:simulation_sparse_add}. We initialized $\boldsymbol{\lambda}$ by considering all possible combinations of $(\lambda_0, \lambda_1 \boldsymbol{1})$ where $\lambda_0, \lambda_1 \in \{10^i: i\in\{-2, -1, 0, 1\}\}$.
	
\begin{figure}
		\caption{\label{fig:mult_starts}
			Error of additive models tuned by Gradient Descent vs. Nelder-Mead. Left: Validation error of models after as the number of initialization points increases. Right: The distribution of validation errors. (Gradient Descent = GD, Nelder-Mead = NM)
		}
		\centering
		\includegraphics[width=0.45\textwidth]{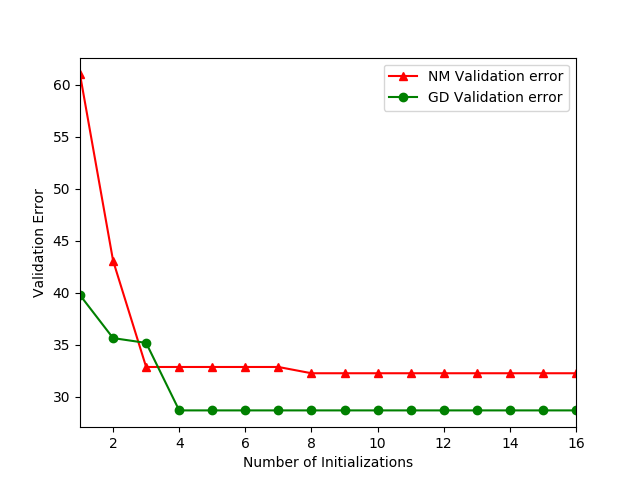}
		\includegraphics[width=0.45\textwidth]{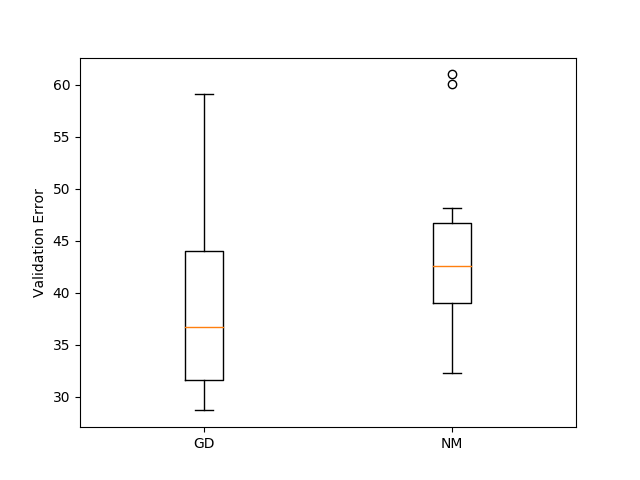}
	\end{figure}

	In Figure \ref{fig:mult_starts} (left), we plot the validation error as the number of initializations increases. The validation errors from both methods plateau quickly. Gradient descent manages to find penalty parameters with lower validation error than Nelder-Mead. Figure \ref{fig:mult_starts} (right) presents the distribution of validation errors resulting from the random initializations. On average, gradient descent finds penalty parameters with lower validation error compared to Nelder-Mead. The plots show that the methods are indeed sensitive to their initialization points. For example, one could run a very coarse grid search on the two-parameter version of the joint optimization problem and use the best penalty parameter values.
	
	\subsection{Additional simulation results}
	
	The simulation results in Section~\ref{sec:results} show that joint optimization problems with many penalty parameters can produce better models than those with only two penalty parameters. One may wonder if this difference is due to the method used to tune the penalty parameters. Here we present results from tuning the two-penalty-parameter joint optimization problems from Sections~\ref{sec:simulation_sparse_add}, \ref{sec:simulation_sgl}, and \ref{sec:simulation_matrix} using gradient descent, Nelder-Mead, and Spearmint. As shown in Table~\ref{tab:two_params}, the performance of these methods are very similar to grid search. Regardless of the method used to tune the two-penalty parameter joint optimization, the resulting models all have higher validation and test error compared to the models from the joint optimization problem with many penalty parameters tuned by gradient descent.
	
	\begin{table}
		\caption {\label{tab:two_params} Two-parameter joint optimization problems for the examples in Section~\ref{sec:results}. Standard errors are given in parentheses. We abbreviated the methods as follows: Gradient Descent = GD, Nelder-Mead = NM, Spearmint = SP, Grid Search = GS}
		\centering
		\begin{tabular}{| l | l | l | l | }
			\hline
			\multicolumn{4}{|c|}{Sparse additive models}\\
			\hline
			& Validation Error & Test Error & \# Solves\\
			\hline
			GD & 23.87 (0.97) & 26.10 (0.86) & 13.07 \\
			\hline
			NM & 28.86 (1.04) & 29.97 (0.96) & 100 \\
			\hline
			SP & 29.18 (1.07) & 30.09 (1.08) & 100 \\
			\hline
			GS & 28.71 (0.97) & 29.42 (0.96) & 100 \\
			\hline
		\end{tabular}
		
		\vspace{0.5cm}
		
		\begin{tabular}{| l | l | l | l | }
			\hline
			\multicolumn{4}{|c|}{Sparse Group Lasso}\\
			\hline
			\multicolumn{4}{|c|}{n=90, p=600, M=30}\\
			\hline
			& Validation Err & Test Err & \# Solves \\
			\hline
			GD & 46.82 (2.21) & 49.33 (1.36)& 21.43\\
			\hline
			NM & 46.37 (2.24) & 48.95 (1.35) & 100 \\
			\hline
			SP &  45.70 (2.32) & 49.35 (1.56) & 100 \\
			\hline
			GS & 47.23 (2.26) & 50.01 (1.40) & 100 \\
			\hline
			\multicolumn{4}{|c|}{n=90, p=900, M=60}\\
			\hline
			& Validation Error & Test Error & \# Solves \\
			\hline
			GD  & 45.71 (2.26) & 50.31 (1.93) & 20.77\\
			\hline
			NM  & 44.95 (2.24) & 50.18 (1.82) & 100  \\
			\hline
			SP  & 49.59 (2.27) & 56.54 (2.14) & 100 \\
			\hline
			GS & 45.70 (2.27) & 51.34 (1.86) & 100 \\
			\hline
			\multicolumn{4}{|c|}{n=90, p=1200, M=100}\\
			\hline
			&  Validation Error & Test Error & \# Solves \\
			\hline
			GD & 50.46 (2.30) & 57.02 (1.94) & 19.80 \\
			\hline
			NM & 49.92 (2.33) & 55.46 (1.89) & 100 \\
			\hline
			SP  & 49.70 (2.26) & 56.51 (2.16) & 100 \\
			\hline
			GS & 50.00 (2.16) & 57.14 (2.18) & 100 \\
			\hline
		\end{tabular}
		
		\vspace{0.5cm}
		
		\begin{tabular}{| l | l | l | l |}
			\hline
			\multicolumn{4}{|c|}{Low-rank Matrix Completion}\\
			\hline
			& Validation Err & Test Err &  Num Solves\\
			\hline
			GD  & 0.70 (0.04) &  0.71 (0.04) & 8.03 (0.79) \\
			\hline
			NM & 0.71 (0.04) & 0.71 (0.04) & 100 \\
			\hline
			SP & 0.73 (0.04) & 0.74 (0.04) & 100\\
			\hline
			GS & 0.71 (0.04) & 0.72 (0.04) & 100\\
			\hline
		\end{tabular}
		
	\end{table}
	
	\subsection{Smoothness of the Validation Loss}
	
	Since our algorithm depends on the validation loss being smooth almost everywhere, a potential concern is that the validation loss may not be differentiable at the solution of the joint optimization problem. We address this concern empirically. Based on the simulation study below, we suspect that the minimizer falls exactly at a knot (where our validation loss is not differentiable with respect to $\boldsymbol{\lambda}$) with measure zero. 
	
	In this simulation we solved a penalized least squares problem with a lasso penalty and tuned the penalty parameter to minimize the loss on a separate validation set. We considered a linear model with 100 covariates. The training and validation sets included 40 and 30 observations, respectively. The response was generated data from the model
	$$
	y = \boldsymbol{X}\boldsymbol{\beta} + \sigma\epsilon
	$$
	where $\boldsymbol{\beta} = (1, 1, 1, 0, ..., 0)$. $\epsilon$ and $\boldsymbol{X}$ were drawn independently from a standard Gaussian distribution. $\sigma$ was chosen so that the signal to noise ratio was 2. For a given $\lambda>0$ our fitted $\boldsymbol{\beta}$ minimized the penalized training criterion
	$$
	\boldsymbol{\hat{\beta}}(\lambda) = \arg\min_{\boldsymbol{\beta}} \| \boldsymbol{y} - \boldsymbol{X}\boldsymbol{\beta} \|_T^2 + \lambda \|\boldsymbol{\beta}\|_1
	$$
	We then chose the $\lambda$-value for which $\boldsymbol{\hat{\beta}}(\lambda)$ minimized the validation error.
	
	In our 500 simulation runs, the penalty parameter that minimized the validation loss was never located at a knot: Using a homotopy solver for the lasso, we were able to find the \emph{exact} knots ($\lambda$-values where variables enter/leave the model), and these points never achieved the minimum value of the validation loss. While this is only one example, and not definitive proof, we believe it is a strong indication that it is unlikely for solutions to occur regularly at knots in penalized problems.
	
	In addition, we believe that the behavior of our procedure is analogous to solving the Lasso via sub-gradient descent. In the Lasso setting, sub-gradient descent with a properly chosen step-size will converge to the solution. In addition, if initialized at a differentiable $\beta$-value (ie. with all non-zero entries), then the lasso objective will be differentiable at all iterates in this procedure with probability one. Admittedly, using the sub-gradient method to solve the lasso has fallen out of favor. The current gold-standard methods, such as generalized gradient descent, give sparse solutions at large enough iterates and achieve faster convergence rates.

\bibliographystyle{agsm}
\bibliography{hillclimbing_nonsmooth_clean_with_appendix}

\end{document}